\documentclass{article} %
\usepackage{dense_preprint,times}

\usepackage{amsmath,amsfonts,bm}

\def\eqref#1{equation~\ref{#1}}

\def\1{\bm{1}}

\DeclareMathAlphabet{\mathsfit}{\encodingdefault}{\sfdefault}{m}{sl}
\SetMathAlphabet{\mathsfit}{bold}{\encodingdefault}{\sfdefault}{bx}{n}

\usepackage{float}
\usepackage{algorithm}
\usepackage{algpseudocode}
\usepackage{hyperref}
\hypersetup{colorlinks=true,linkcolor=blue,citecolor=blue,urlcolor=blue}
\usepackage{url}
\usepackage{booktabs}
\usepackage{colortbl}
\usepackage{graphicx}
\usepackage{wrapfig}
\usepackage{tcolorbox}
\usepackage{iftex}
\usepackage{needspace}
\tcbuselibrary{breakable,skins}

\definecolor{promptinterface}{HTML}{355D91}
\definecolor{promptbaseline}{HTML}{79509B}
\definecolor{promptbuild}{HTML}{287D78}
\definecolor{promptreconcile}{HTML}{A16A24}

\newcommand{\promptbodyfont}{%
  \small\ttfamily
  \ifPDFTeX\fontencoding{T1}\fontfamily{lmtt}\selectfont\fi
}

\newtcolorbox[auto counter]{promptbox}[3]{
  enhanced jigsaw,
  breakable,
  shrink break goal=1pt,
  colback=#1!2!white,
  colframe=#1,
  colbacktitle=#1!10!white,
  coltitle=#1!45!black,
  title={Prompt P\thetcbcounter\quad #2},
  label={#3},
  title after break={Prompt P\thetcbcounter\quad #2 (continued)},
  fonttitle=\small\sffamily\bfseries,
  fontupper=\promptbodyfont,
  boxrule=0.65pt,
  arc=1.5mm,
  left=2.8mm,right=2.8mm,top=2mm,bottom=2mm,
  before skip=9pt,after skip=10pt,
  pad at break=2mm,
  lines before break=5,
  before upper={\raggedright\setlength{\parindent}{0pt}\setlength{\parskip}{4pt}}
}

\newcommand{\promptrole}[1]{%
  \par\smallskip{\normalfont\sffamily\bfseries\footnotesize #1}\par\nobreak
}
\newcommand{\promptomission}{%
  {\normalfont\itshape[\ldots{} excerpt omitted \ldots{}]}%
}

\newcommand{\PromptRecipient}{%
\begin{promptbox}{promptinterface}{Feedback for a fresh execution}{prompt:recipient}
\promptrole{Recipient-side template (complete)}
\# Previous Attempt Analysis

This rerun starts in a fresh isolated environment. Do not assume that files, processes, temporary directories, or session state from the previous attempt still exist. Paths and artifacts mentioned below are historical evidence only. Recreate every required deliverable in the current environment. Reuse the approach or content, not previous filesystem state.

Below is an annotated analysis of a previous attempt at this task. Use it as evidence about what to preserve, correct, or recreate in this attempt.

{\setlength{\parskip}{1pt}
\textless{}report\textgreater{}\par
\{content\}\par
\textless{}/report\textgreater{}\par
}
\end{promptbox}%
}

\newcommand{\PromptAdvisor}{%
\begin{promptbox}{promptbaseline}{Task-only Advisor}{prompt:advisor}
\promptrole{System message}
You are a task-only advisor for an AI agent before it begins a task. You receive only the original task instruction. You have not observed any previous execution, reasoning, action, tool call, tool result, test output, reward, verifier feedback, reference answer, or hidden state. Base your advice only on general knowledge and the task instruction. Never claim or imply that a particular event, error, file, command result, or failure has already occurred. Do not fabricate observations.

\promptrole{User message (requested-advisory excerpt)}
{\setlength{\parskip}{1pt}
Provide concise, actionable guidance for an agent that has not attempted the task yet. Use these Markdown sections:\par
1. \textasciigrave{}\#\#\# Recommended Strategy\textasciigrave{} \textemdash{} a practical sequence of steps;\par
2. \textasciigrave{}\#\#\# Likely Risks\textasciigrave{} \textemdash{} pitfalls inferable from the instruction alone;\par
3. \textasciigrave{}\#\#\# Verification Checklist\textasciigrave{} \textemdash{} checks the agent should perform before finishing.\par
\par
Do not discuss a previous attempt and do not claim access to execution evidence. Return only the advisory Markdown.\par
}

\promptrole{Recipient-side template (complete)}
\# Task-Only Advisory

The following advisory was generated before execution from the original task instruction alone. The advisor did not observe any previous attempt or outcome.

{\setlength{\parskip}{1pt}
\textless{}advisory\textgreater{}\par
\{content\}\par
\textless{}/advisory\textgreater{}\par
}
\end{promptbox}%
}

\newcommand{\PromptAllAtOnce}{%
\begin{promptbox}{promptbaseline}{All-at-once trajectory scoring}{prompt:all-at-once}
\promptrole{System message}
You are an expert evaluator of AI agent trajectories. You are given a task description and an agent's COMPLETE execution trajectory as an OA (observation-action) sequence. Score EVERY action in a single pass. Because you see the whole trajectory and all of your own verdicts at once, keep the labels mutually consistent: judge each action by what it contributed to the trajectory as a whole, not in isolation. An action that looked reasonable locally but was made redundant, undone, or invalidated by what followed should be labelled accordingly.

\promptrole{User message: scoring criteria}
{\setlength{\parskip}{1pt}
label ranges from -2 to 0, where:\par
-  0: an effective action that helps complete the task\par
- -1: an ineffective or failed action, but not severe enough to fail the task\par
- -2: a severe, fatal error that may directly cause the task to fail\par
}

\promptrole{User message: required-output excerpt}
{\setlength{\parskip}{1pt}
Score all of the actions listed above in ONE pass \textemdash{} exactly one entry per action\_index, no more and no less. Output ONLY a JSON array, nothing else:\par
[\{"action\_index": \textless{}int\textgreater{}, "reasoning": "\textless{}str\textgreater{}", "label": \textless{}int\textgreater{}\}, ...]\par
Note: within each object, output reasoning first (the complete analysis), then give label based on that reasoning. Keep each reasoning to a couple of sentences so every action fits in one response.\par
}
\end{promptbox}%
}

\newcommand{\PromptStepByStep}{%
\begin{promptbox}{promptbaseline}{Step-by-step decision scoring}{prompt:step-by-step}
\promptrole{System message}
You are an expert evaluator of AI agent trajectories, scoring one action of an AI agent in real time. You are given a task description and the trajectory PREFIX up to and including the target action \textemdash{} exactly what the agent itself had seen when it decided to act. Everything that happened afterwards is deliberately hidden from you: you do NOT know whether the action succeeded. Judge the DECISION, not the outcome. Ask: given only the information available at this point, was this the right move? A well-grounded action that later fails for reasons nobody could foresee is still a good decision; a guess that happens to work is still a bad one. Concretely, weigh whether the action: follows from evidence already in the prefix rather than from an unverified assumption; is the most direct available next step toward the task rather than a detour or a repeat of work already done; is well-formed and correctly parameterized given what is known; and gathers the information it needs before committing to an irreversible change.

\promptrole{User message: scoring criteria}
{\setlength{\parskip}{1pt}
label ranges from -2 to 0, where:\par
-  0: a well-justified decision given what was known at this point\par
- -1: a poorly justified decision \textemdash{} an unverified assumption, an avoidable detour, or a repeat of work already done \textemdash{} but not fatal\par
- -2: a severely misguided decision that a careful agent could have avoided with the information already available, and that may derail the task\par
}

\promptrole{User message: evidence and output excerpt}
{\setlength{\parskip}{1pt}
known\_basis: one sentence naming the concrete evidence ALREADY IN THE PREFIX that justifies this action (cite the index, e.g. "the file list from index 4"), or \textemdash{} if there is none \textemdash{} what the agent would have needed to check first.\par
Do NOT speculate about what happens next; you have not been shown it.\par
Output ONLY JSON, nothing else: \{"action\_index": \textless{}int\textgreater{}, "known\_basis": "\textless{}str\textgreater{}", "reasoning": "\textless{}str\textgreater{}", "label": \textless{}int\textgreater{}\}\par
Note: output known\_basis and reasoning before label, so the conclusion follows from the evidence.\par
}
\end{promptbox}%
}

\newcommand{\PromptBoundary}{%
\begin{promptbox}{promptbuild}{DENSE: subtask boundaries}{prompt:boundary}
\promptrole{System message}
You are an expert at structuring AI agent trajectories, reconstructing an execution trajectory into a hierarchical subtask tree in a [bottom-up] manner. You will see the root task description, the current tree-building level, and a [contiguous, numbered] view of actions/sub-phases (a queue). Your task is to decide: from the head of the queue up to which element exactly forms one [complete and self-contained sub-phase/subtask] \textemdash{} i.e. those elements together accomplish a relatively independent thing, while the element immediately following clearly opens a new purpose. Like identifying function boundaries, split only at natural semantic breaks; prefer to wait for more actions to enter the view rather than forcibly merging elements that clearly span multiple different purposes into one sub-phase. [Prefer split] when the queue transitions from exploration/debugging to committing irreversible state (overwriting files, opening a DB that may checkpoint/delete WAL, deleting artifacts, installing packages, starting long-lived processes), or from discovery to recipe execution \textemdash{} do not glue the trial-and-error stretch and the final commit into one oversized sub-phase.

\promptrole{User message: required output}
{\setlength{\parskip}{1pt}
Output ONLY JSON, nothing else: \{"reasoning": "\textless{}first analyze the queue view: which elements belong to the same thing, where it breaks, or why it cannot break yet\textgreater{}", "action\_index": \textless{}int\textgreater{}\}\par
Rules for action\_index (the number is the [ordinal] labeled on each line of the view, increasing consecutively from the head element's ordinal, NOT a 0-based array index):\par
- output some ordinal i (queue-head ordinal \textless{}= i \textless{}= queue-tail ordinal): means the elements from the queue head up to ordinal i [together form one complete sub-phase], and ordinal i+1 onward clearly opens a new purpose. These elements will be aggregated into an upper-level node.\par
- output 0: means the [single] element at the queue head is itself already a complete sub-phase; aggregate just it alone.\par
- output -1: means the elements currently in view are not yet enough to form a complete sub-phase; wait for more actions to enter the queue before deciding.\par
Note: reasoning must come before action\_index, so the conclusion is driven by the analysis.\par
}
\end{promptbox}%
}

\newcommand{\PromptLocalScore}{%
\begin{promptbox}{promptbuild}{DENSE: local action scoring}{prompt:local-score}
\promptrole{System message}
You are an expert evaluator of AI agent trajectories, scoring an agent trajectory action by action inside a [just-formed sub-phase]. You will see: the root task (instruction), an XML mini-tree whose only subtask is the current sub-phase (focus="current-subtask"), and one [target action] marked with focus="TARGET". Your scoring [MUST] be grounded in the root instruction and this sub-phase's local goal:

1. Infer the sub-phase's local goal from the actions it contains (subtitle/summary may be empty at this stage);

2. Judge whether the target action advances that local goal, is ineffective/repetitive/a detour, or damages progress; necessary exploration, debugging, and error-correction are valuable and should not be judged ineffective merely because they did not immediately produce a result;

3. Align with the root task's final goal: even if locally reasonable, reflect clear deviation or harm faithfully;

4. Combine with the environment feedback (observation) immediately following the target action. Always keep your scoring basis on [the root task requirements + this sub-phase's local role].

\promptrole{User message: scoring-rubric excerpt}
{\setlength{\parskip}{1pt}
label ranges from -2 to 0, where:\par
-  0: an effective action that helps complete the task, e.g. correctly advancing task progress, successfully building the environment, achieving a milestone, or advancing the local goal of its subtask / the final goal of the root task\par
- -1: an ineffective or failed action (repetition, detour, no progress, a failed tool call, or no help to the task) that leaves the task no worse off \textemdash{} the agent can still recover from here. A failed attempt that is later corrected by other means belongs to -1, not -2\par
- -2: an action that actively damages the task outcome, i.e. the delivered result is wrong (or is wrongly believed to be right) because of it. Typical cases:\par
}
\promptomission

\promptrole{User message: required output}
{\setlength{\parskip}{1pt}
Output ONLY JSON, nothing else: \{"reasoning": "\textless{}scoring reasoning\textgreater{}", "label": \textless{}int\textgreater{}\}\par
Note: reasoning must come before label, so the conclusion is driven by the reasoning.\par
}
\end{promptbox}%
}

\newcommand{\PromptSummary}{%
\begin{promptbox}{promptbuild}{DENSE: evidence-preserving summaries}{prompt:summary}
\promptrole{System message}
You are an expert evaluator of AI agent trajectories. A group of consecutive actions/sub-phases has just been identified as a higher-level sub-phase. Give it a short subtitle, and based on the sequence of actions it contains and their execution results (and per-action labels / score reasoning when present), distill the key information that an agent rerunning this task later can directly reuse. Focus on: which files/artifacts were finally produced (and their paths), what cleanup was done to the environment when the sub-phase ended (restore/cleanup/leftovers), the key values and mappings extracted or computed, critical ordering constraints, and whether there are any unresolved issues. In addition, specifically distill the load-bearing key implementation mechanisms of this sub-phase \textemdash{} the required step order, irreversible state changes, correct API/parameter mappings, environment prerequisites, and (only when present) concurrency/async/daemon design whose omission on rerun would cause functional regression.

 [Shortcut rule] The summary is what a rerun agent reads INSTEAD of the step-by-step detail. Write it as the [shortest path that actually worked], not as a chronicle of the search. If the agent tried tools/approaches A, B, C and only D worked, the summary says "used D to do X" \textemdash{} it does NOT narrate trying A, B, C first. [Exception \textemdash{} never shortcut away verification]: if the subtask ran a check, test, or validation that substantiates its result, that step MUST stay in the summary (state what was verified and how), because a rerun agent that skips verification will silently ship a wrong result. Failed attempts are not discarded: put them in \textasciigrave{}dead\_ends\textasciigrave{} so the rerun agent can avoid them, but keep them out of the summary narrative.

[Evidence-grounded values] Put a value into \textasciigrave{}key\_values\textasciigrave{} only when a later action in this sub-phase verifies it (re-read file, query/test, cross-check against the instruction). If the agent only computed or wrote a result without verification, put \textasciigrave{}unverified result: ...\textasciigrave{} in \textasciigrave{}open\_issues\textasciigrave{} \textemdash{} do NOT state it as a reusable fact in \textasciigrave{}key\_values\textasciigrave{}.

[Critical ordering] Extract must-do-before constraints for irreversible/stateful steps (overwrite, DB open that may checkpoint, delete, install, start daemon) into \textasciigrave{}critical\_order\textasciigrave{} and also mention them in the summary when they are load-bearing.

[Mine failures] Scan execution results (and negative labels when present) for reusable dead ends (command not found, traceback, permission errors, 404, incompatible versions). Put those in \textasciigrave{}dead\_ends\textasciigrave{}; leave one-off typos the agent immediately corrected out.

[Parent reconcile] When children are already sub-phases with conflicting \textasciigrave{}key\_values\textasciigrave{}, keep the later verified value; mark superseded values as dead ends \textemdash{} never present two contradictory authoritative parameter sets.
\end{promptbox}%
}

\newcommand{\PromptTermination}{%
\begin{promptbox}{promptbuild}{DENSE: hierarchy termination}{prompt:termination}
\promptrole{System message}
You are an expert at structuring AI agent trajectories, deciding whether a hierarchical subtask tree is already [fully built]. There is currently a set of candidate top-level nodes, plus the root task description. Only when [every one of these nodes is exactly one top-level major phase of the root task] (coarse-grained enough, with clear boundaries between them, together fully covering the whole task), and the number of nodes is already few enough (a handful of trunk phases, not a long list of fine-grained nodes that could still be merged), should you judge the tree building complete and mount them all directly under the root node. If these nodes are still too many, uneven in granularity, or several of them clearly could be further aggregated into larger phases, you [MUST] judge it not yet complete and aggregate one more level upward. [Important] Do NOT judge it complete just because 'these nodes can all technically be mounted under the root' \textemdash{} almost any node can be mounted under the root, that is not the criterion; the criterion is whether they are already the [coarsest-grained, no-longer-naturally-aggregatable] top-level phase division. If candidate summaries still mix unresolved exploration with the final recipe, or present conflicting key values that a further aggregation could reconcile, lean toward not-yet-complete. First [node by node] briefly state whether it holds up as a direct sub-phase of the root task and whether it could still be aggregated with adjacent nodes, then give the overall conclusion.

\promptrole{User message: required output}
{\setlength{\parskip}{1pt}
Output ONLY JSON, nothing else: \{"reasoning": "\textless{}node-by-node analysis of whether each is already a top-level major phase of the root task, whether it can be aggregated further, and whether the overall count and granularity are appropriate\textgreater{}", "can\_mount\_all": \textless{}true|false\textgreater{}\}\par
Note: reasoning must come before can\_mount\_all. Lean strict: as long as further aggregation is clearly still possible, output false.\par
}
\end{promptbox}%
}

\newcommand{\PromptCleaner}{%
\begin{promptbox}{promptreconcile}{DENSE: contiguous shortcut compression}{prompt:cleaner}
\promptrole{System message}
You are the path cleaner for Shortcut-Tree, an evidence-preserving representation of an AI-agent execution trace. You receive the ordered direct children of exactly one parent subtask. Partition those siblings without reordering them. Replace a contiguous try/fail/recover stretch with one shortcut node that records the reusable dead end, the shortest working path, and the observed result. If every attempt failed, say so and retain an open issue. Do not replay the search chronicle.

Numeric action labels are hints about where to inspect, never proof of success. Success requires cited observation, verification, or an already-criticized child subtask. Paths and artifacts are historical evidence from an earlier attempt; never imply that they exist in the rerun environment. Never invent a path, value, artifact, result, or verification without a supplied source/evidence ID. Retain decisive correct or incorrect actions and any key action identified in the input. Output a strict, complete, contiguous partition.

\promptrole{User message: required output}
{\setlength{\parskip}{1pt}
Output ONLY JSON:\par
\{\par
\hspace*{1.0em}"groups": [\par
\hspace*{2.0em}\{\par
\hspace*{3.0em}"mode": "keep|shortcut",\par
\hspace*{3.0em}"source\_node\_ids": ["direct-child-id", "..."],\par
\hspace*{3.0em}"title": "short factual title",\par
\hspace*{3.0em}"dead\_end": "reusable failed approach and observed reason, or none",\par
\hspace*{3.0em}"working\_path": "shortest evidence-backed path that worked, or empty",\par
\hspace*{3.0em}"outcome": "succeeded|failed|partial|unknown",\par
\hspace*{3.0em}"open\_issue": "unresolved requirement, or empty",\par
\hspace*{3.0em}"evidence\_node\_ids": ["supplied-id"],\par
\hspace*{3.0em}"key\_action\_ids": ["supplied-action-node-id"]\par
\hspace*{2.0em}\}\par
\hspace*{1.0em}]\par
\}\par
Rules: a keep group contains exactly one source; a shortcut contains at least\par
two contiguous sources. Every direct child must occur exactly once and in the\par
original order.\par
}
\end{promptbox}%
}

\newcommand{\PromptCritic}{%
\begin{promptbox}{promptreconcile}{DENSE: evidence-based issue reconciliation}{prompt:critic}
\promptrole{System message}
You are the subtree critic for Shortcut-Tree. Judge only the cleaned subtree and its cited evidence. Decide whether the local theme is internally coherent and whether its subtask was actually completed. Identify unresolved ordinary issues and fatal issues. A confident statement or a numeric action label is not evidence. Success requires an observation, verification, or a coherent completed child verdict. Historical artifacts do not persist into a rerun environment.

You may resolve an issue inherited from a child only when a later cleaned node contains concrete recovery evidence. For every resolved issue, return non-empty resolution\_evidence containing supplied IDs. Select key actions by ID; never rewrite their original content. Do not output a broken flag: status is derived deterministically after validation.

\promptrole{User message: completion scope (also used by the cleaner)}
\#\# Completion scope No reward or verifier output was supplied. A complete verdict means only that the supplied trajectory evidence is internally coherent and appears to satisfy the task. It is not verifier-confirmed correctness.

\promptrole{User message: required output}
{\setlength{\parskip}{1pt}
Output ONLY JSON:\par
\{\par
\hspace*{1.0em}"coherence": "coherent|inconsistent|insufficient\_evidence",\par
\hspace*{1.0em}"completion": "complete|incomplete|failed|unknown",\par
\hspace*{1.0em}"summary": "shortest evidence-grounded account of the final state",\par
\hspace*{1.0em}"open\_issues": [\par
\hspace*{2.0em}\{"issue": "description", "evidence\_node\_ids": ["supplied-id"]\}\par
\hspace*{1.0em}],\par
\hspace*{1.0em}"fatal\_issues": [\par
\hspace*{2.0em}\{"issue": "description", "evidence\_node\_ids": ["supplied-id"]\}\par
\hspace*{1.0em}],\par
\hspace*{1.0em}"resolved\_issue\_ids": ["inherited-issue-id"],\par
\hspace*{1.0em}"resolution\_evidence": \{"inherited-issue-id": ["supplied-id"]\},\par
\hspace*{1.0em}"key\_action\_ids": ["supplied-action-node-id"]\par
\}\par
}
\end{promptbox}%
}

\usepackage{fvextra}
\usepackage{array}
\usepackage{capt-of}
\usepackage{placeins}
\newcolumntype{L}[1]{>{\raggedright\arraybackslash}p{#1}}
\usetikzlibrary{arrows.meta,positioning}
\definecolor{caseresolved}{HTML}{287D78}
\definecolor{caseopen}{HTML}{A16A24}
\definecolor{caseevidence}{HTML}{355D91}
\newtcolorbox{casebox}[1]{
  enhanced jigsaw,breakable,
  title after break={#1 (continued)},
  lines before break=4,
  colback=caseevidence!2!white,colframe=caseevidence!65!black,
  colbacktitle=caseevidence!10!white,coltitle=black,
  title={#1},fonttitle=\small\sffamily\bfseries,
  boxrule=.6pt,arc=1mm,left=2mm,right=2mm,top=2mm,bottom=2mm
}
\newcommand{\caseVerbatim}[1]{%
  \IfFileExists{#1}{%
    \VerbatimInput[breaklines=true,breakanywhere=true,breakautoindent=false,
      fontsize=\footnotesize,baselinestretch=1.05]{#1}%
  }{%
    \VerbatimInput[breaklines=true,breakanywhere=true,breakautoindent=false,
      fontsize=\footnotesize,baselinestretch=1.05]{../#1}%
  }%
}

\newcommand{\caseQuoteAdvisor}{Format the discovered matches (one per line) and write them to /app/recovered\_passwords.txt.}
\newcommand{\caseQuoteGlobal}{Utilizing byte offsets to locate strings is a high-quality forensic approach that helps pinpoint the exact location of the target data.}
\newcommand{\caseQuotePrefix}{Using grep -boa to find offsets is the correct forensic step to then examine the raw bytes around that location and reconstruct the full 23-character password.}
\newcommand{\caseQuoteDense}{The identified password '8XDP5Q2RT9ZK7VB3BV4WW54' must be written to /app/recovered\_passwords.txt.}

\graphicspath{{figures/}}

\definecolor{refitheader}{RGB}{244,247,250}
\definecolor{refitmodel}{RGB}{228,236,245}
\definecolor{refitours}{RGB}{239,246,252}
\definecolor{refitsts}{RGB}{248,242,231}
\definecolor{refitpriv}{RGB}{247,247,247}

\title{DENSE: Distilling Agent Trajectories into\\
Evidence-Grounded Shortcut Trees for Self-Refinement}

\author{Siyuan Liu\textsuperscript{1,2}, Fan Yu\textsuperscript{1,2},
Dongyu Ru\textsuperscript{2}, Yizhu Liu\textsuperscript{2} \\
\textbf{Yifan Yang\textsuperscript{2}, Xuezhi Cao\textsuperscript{2},
Xunliang Cai\textsuperscript{2}, Yixin Cao\textsuperscript{1}} \\
\textsuperscript{1}\,Fudan University \qquad \textsuperscript{2}\,Meituan Longcat Team
}

\hypersetup{
  pdftitle={DENSE: Distilling Agent Trajectories into Evidence-Grounded Shortcut Trees for Self-Refinement},
  pdfauthor={Siyuan Liu, Fan Yu, Dongyu Ru, Yizhu Liu, Yifan Yang, Xuezhi Cao, Xunliang Cai, Yixin Cao}
}
\date{}

\begin{document}

\maketitle

\begin{abstract}
Online agent deployments accumulate execution trajectories at massive
scale and behavioral diversity, for which predefined annotation criteria hardly exist.
Extracting useful evidence therefore demands costly manual annotation or verifier signals that
fail to scale, leaving valuable evidence buried among redundant, incomplete,
and failed executions. This raises a question:
without post-execution rewards or correctness labels, how can reusable experience be distilled 
from the trajectories themselves?
To address this challenge, we introduce \textbf{DENSE} (Distilling
Evidence from Nested Subtask Executions), which organizes trajectory-derived
evidence into \emph{nested shortcut trees}. By consolidating redundant attempts, identifying resolved
subtasks, and retaining useful steps alongside outstanding requirements, DENSE transforms noisy
 execution traces into structured and reusable
task-solving feedback. To evaluate whether such feedback helps agents retry
the same task, we design \textbf{REFIT}, which measures success-rate changes between
the initial attempt and feedback-guided retries. Among feedback methods without external outcome 
supervision, DENSE achieves the highest strict pass rate
across four agent models on Terminal-Bench 2.1, improving over initial attempts
by \textbf{7.12--21.81 percentage points} with \mbox{\textbf{19.0--43.6\%}} fewer agent tokens on retries.
In addition, on hard tasks DENSE consistently outperforms self-reflection in cumulative pass rate 
across multiple feedback iterations on all four models, demonstrating its strong potential for 
continual agent self-improvement.
\end{abstract}

\edef\refitRestoreSpacing{%
  \noexpand\setlength{\noexpand\parskip}{\the\parskip}%
  \noexpand\setlength{\noexpand\textfloatsep}{\the\textfloatsep}%
  \noexpand\setlength{\noexpand\intextsep}{\the\intextsep}%
  \noexpand\setlength{\noexpand\floatsep}{\the\floatsep}%
  \noexpand\setlength{\noexpand\abovecaptionskip}{\the\abovecaptionskip}%
}
\setlength{\parskip}{4.5pt}
\setlength{\textfloatsep}{14pt plus 2pt minus 2pt}
\setlength{\intextsep}{10pt plus 2pt minus 2pt}
\setlength{\floatsep}{10pt plus 2pt minus 2pt}
\setlength{\abovecaptionskip}{8pt}

\par\vspace{-3pt}
\section{Introduction}
\vspace{-2pt}
\label{sec:intro}

\begin{wrapfigure}{r}{0.48\textwidth}
  \vspace{-\intextsep}
  \centering
  \includegraphics[trim=0bp 0bp 178bp 8bp,clip,width=\linewidth]{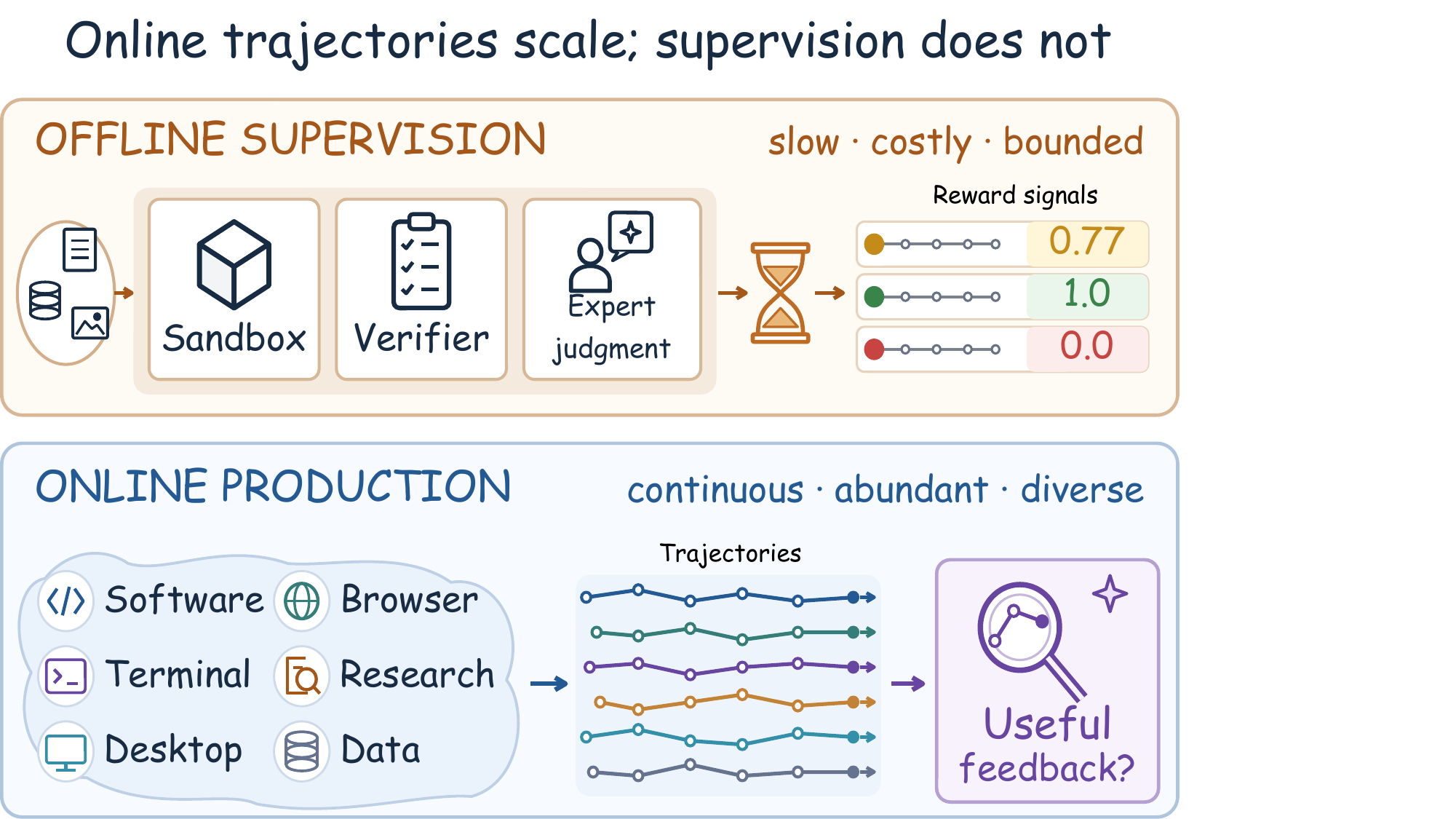}
  \small
  \caption{Execution traces accumulate, while external supervision is costly
  and bounded. Can agents reuse experience from these traces without external
  rewards or correctness labels?}
  \label{fig:supervision-gap}
  \vspace{-6pt}
\end{wrapfigure}

Large language model agents increasingly operate in software engineering
\citep{yang2024sweagent}, web interaction \citep{zhou2023webarena}, deep research
\citep{tongyi2025deepresearch}, and desktop automation \citep{xie2024osworld}.
In offline evaluation, curated tasks can be assessed using task-specific
verifiers or human expert annotations. Online deployment continuously generates
long traces of actions, observations, failed attempts, and recovery across
diverse tasks; providing comparable supervision for these traces is costly and
difficult to scale (Figure~\ref{fig:supervision-gap}). This gap raises a question: \textbf{without
post-execution rewards or correctness labels, what useful experience can we extract from the trajectories
themselves?}

Prior work demonstrates that trajectory experience can improve subsequent
execution. Reflexion uses evaluator rewards to guide verbal reflection
\citep{shinn2023reflexion}; ExpeL extracts lessons by contrasting successful and
failed attempts \citep{zhao2024expel}; and DoVer tests failure hypotheses through
interventions and outcome verification \citep{ma2025dover}. These approaches
improve task performance, but depend on outcome signals to guide reflection,
experience selection, or diagnosis. However, how to systematically extract
reusable experience from \emph{unlabeled online trajectories}, without
post-execution rewards, verifier outputs, or human correctness annotations,
remains underexplored.

Our central insight is that \textbf{trajectories still contain reusable evidence,
even without outcome labels}. For example, a tool call may fail because of a
missing argument, then return the requested data after correction. This sequence
supports reusing the corrected call, even when overall task completion is
uncertain and the requested analysis remains unverified. The challenge is to
connect useful steps to supporting observations and outstanding requirements.
We call this process \emph{trajectory evidence distillation}: constructing
feedback for subsequent execution using only the task description, recorded
observations, and its execution trace, without post-execution
rewards or correctness labels.

To efficiently distill this evidence into reusable feedback, we propose \textbf{DENSE}
(Distilling Evidence from Nested Subtask Executions), which constructs
\emph{evidence-grounded nested shortcut
trees}. Each shortcut records attempted approaches, useful steps, supporting
observations, and unfinished requirements within a subtask.
DENSE groups actions into nested subtasks and condenses repeated attempts.
When combining subtasks, it updates unresolved issues if later observations show
that they have been resolved. It then summarizes completed subtasks and retains
more detail for unfinished ones, preserving the actions needed to recreate earlier
progress.

To evaluate the value of experience extracted from online trajectories by different
methods, we introduce \textbf{REFIT}, which compares feedback constructed from
the same initial trajectory (run0) without access to post-execution outcome
information. The same agent then makes a separate attempt (run1) with each method's
feedback, starting from the initial environment state and a fresh model context.
Only feedback carries information between attempts.
We compare strict pass rates and changes from run0 under matched task, tool,
and execution-budget conditions. Following critique evaluation through subsequent
answers \citep{tang2025realcritic}, we assess feedback by task outcomes, keeping
verifier results separate from feedback generation.

Our contributions are:
\vspace{-2pt}
\begin{enumerate}
  \setlength{\itemsep}{1pt}
  \setlength{\parsep}{0pt}
  \item \textbf{The DENSE method.} We introduce evidence-grounded nested shortcut
  trees that preserve useful steps and unfinished requirements, using subtask
  structure to track issue resolution and prioritize feedback detail according to
  each subtask's remaining requirements.

  \item \textbf{The REFIT framework.} We introduce a protocol that compares
  feedback methods using shared initial trajectories, reset environments and
  contexts, and separate outcome evaluation, while excluding post-execution
  outcome signals from feedback generation.

  \item \textbf{Systematic evaluation and mechanism analysis.} Using REFIT, we
  compare DENSE with alternative feedback methods on Terminal-Bench 2.1
  \citep{merrill2026terminalbench}. DENSE achieves the highest strict pass rate
  among the tested non-privileged methods\footnote{By privileged feedback, we
  mean exposing to the model post-hoc reward information and verifier outputs
  that are otherwise hidden during execution.} across four agent models, improving over
  run0 by \textbf{7.12--21.81 pp} and reducing observed agent token use in run1 by
  \mbox{\textbf{19.0--43.6\%}}. Ablations and case studies probe the mechanisms;
  an exploratory hard-task extension shows higher cumulative pass rates than
  Self-reflection after three feedback iterations, suggesting DENSE's potential to support
  agent refinement over multiple rounds.
\end{enumerate}

\par\vspace{-3pt}
\section{DENSE: Evidence-Grounded Shortcut Trees}
\vspace{-2pt}
\label{sec:method}

DENSE organizes scattered actions and observations into reusable task experience
in two stages. It first identifies semantic boundaries in ordered queues to
build a nested subtask tree. It then compresses attempts into shortcuts,
reconciles issues across levels, and expands feedback according to issue state.
The following sections explain these steps through one CSV report example.

\par\vspace{-2pt}
\subsection{Problem Setting and Overview}
\vspace{-3pt}

\begin{figure}[t]
  \vspace{-4pt}
  \centering
  \includegraphics[trim=2bp 4bp 2bp 4bp,clip,width=0.98\linewidth]{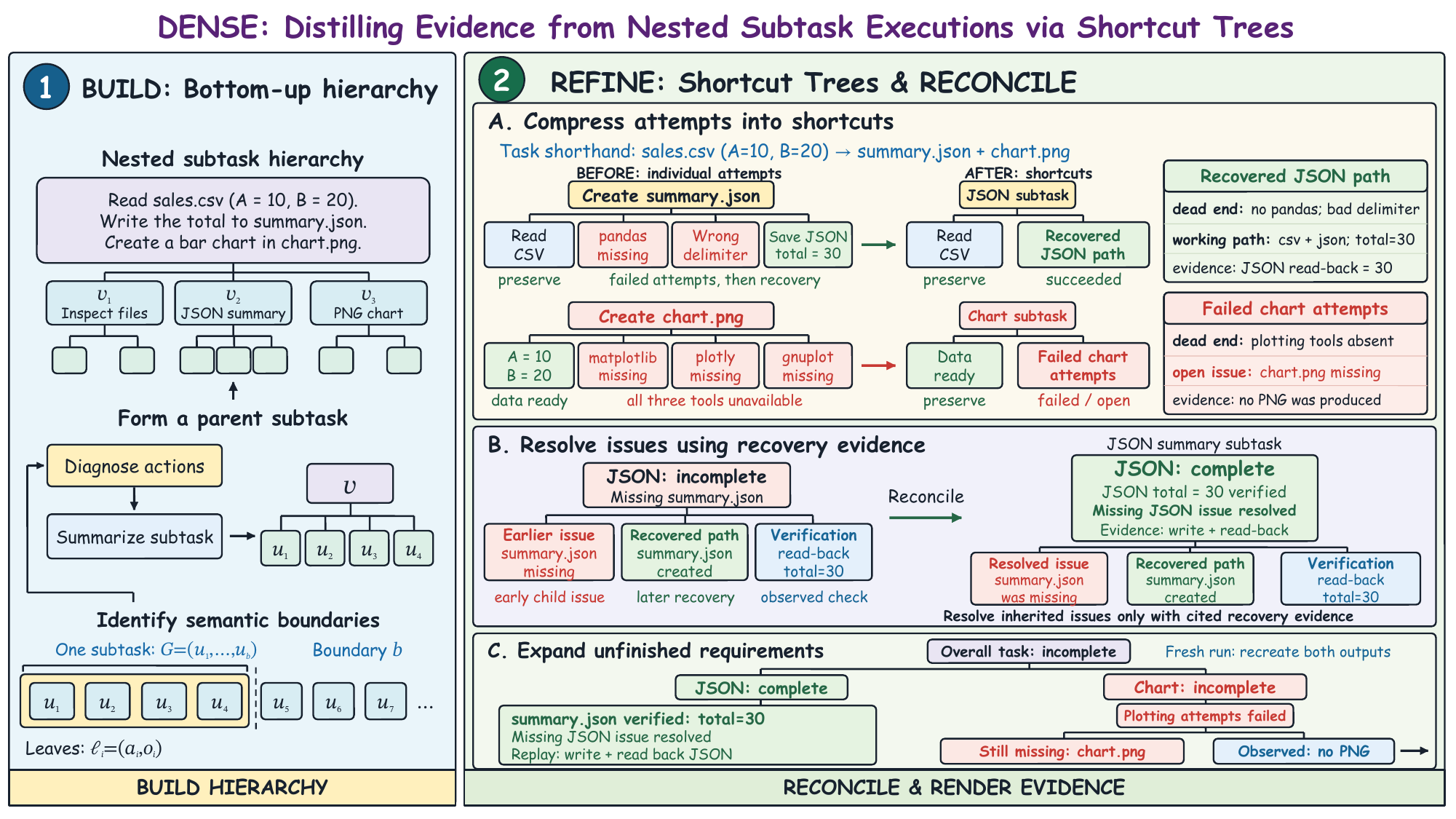}
  \caption{DENSE has two stages. Stage 1 identifies a semantic boundary for a
  consecutive prefix of queue $Q_k$, diagnoses its actions, summarizes a parent
  node, and places it in $Q_{k+1}$ to build the hierarchy. Stage 2 has three
  steps: A compresses sibling attempts into shortcuts retaining failed attempts,
  working paths, and evidence; B reconciles inherited issues using later recovery
  evidence; C summarizes completed branches and expands unfinished requirements
  with their evidence. The constructed CSV report example has a completed JSON
  summary and an unfinished PNG chart.}
  \label{fig:dense-pipeline}
  \vspace{-5pt}
\end{figure}

Given task $x$ and a finished attempt's trajectory
$\tau=(a_1,o_1,\ldots,a_T,o_T)$, DENSE produces bounded feedback distinguishing
useful progress, recovered local errors, and unfinished requirements
(Figure~\ref{fig:dense-pipeline}). It reads only the task and visible trajectory;
Section~\ref{sec:refit} defines the information boundary and reset conditions.

\par\vspace{-2pt}
\subsection{Building Nested Subtasks}
\vspace{-3pt}

Local outcomes must be interpreted within their goals. DENSE pairs actions and
observations as leaves $\ell_i=(a_i,o_i)$ in a temporally ordered queue $Q_k$.
The analysis model identifies a prefix $G=(u_1,\ldots,u_b)$ serving one objective,
expanding its visible range when evidence is insufficient. After selecting
boundary $b$, it groups these nodes under parent $v$, diagnoses their actions,
summarizes the subtask, and places $v$ in $Q_{k+1}$. Repetition builds the root
hierarchy. Each level preserves temporal order and source coverage, allowing
local outcomes to be reinterpreted within higher-level goals.

In Figure~\ref{fig:dense-pipeline}, the task is to read \texttt{sales.csv} and
produce a total in \texttt{summary.json} and a bar chart in \texttt{chart.png}.
After encountering missing \texttt{pandas} and an incorrect delimiter, the JSON
branch uses the standard library to write the summary and read back a total of
30. The plotting branch tries three unavailable tools and produces no PNG.
Separate subtasks represent both the completed summary and the unfinished chart.

\par\vspace{-2pt}
\subsection{Compressing Attempts with Their Evidence}
\vspace{-3pt}

DENSE processes the tree bottom-up, fusing consecutive attempts under one parent
into a shortcut. Each shortcut retains failed approaches, a compact working path
supported by observations, the final local outcome, and source references.
Unresolved shortcuts retain unmet requirements and relevant observations.
Fusion never crosses parent boundaries or reorders sources.

In the CSV example, the JSON shortcut compresses the missing-\texttt{pandas},
incorrect-delimiter, and successful-write attempts. It records the first two as
failed approaches and retains the working \texttt{csv}/\texttt{json} path with
the observation confirming a total of 30. The plotting shortcut instead records
the three unavailable tools and the missing \texttt{chart.png}. The same
structure thus supplies historically supported paths and experience that helps
the next attempt avoid repeating unproductive exploration.

\par\vspace{-2pt}
\subsection{Reconciling Issues through Recovery Evidence}
\vspace{-3pt}

An early issue may be repaired by actions in a later subtask, so a parent must
revisit issues inherited from its children. For each compressed subtree, the
analysis model produces a final-state summary and any unresolved or fatal issues.
The parent combines child summaries, issues, and recovery evidence:
\begin{equation}
  (\mathcal T_v^\star,\sigma_v,\mathcal I_v)
  =\operatorname{Reconcile}\!\left(
    \widetilde{\mathcal T}_v,
    \{(\sigma_u,\mathcal I_u)\}_{u\in\operatorname{child}(v)}
  \right).
  \label{eq:dense-reconciliation}
\end{equation}
Here $\widetilde{\mathcal T}_v$ is the compressed local subtree, $\sigma_v$ its
state summary, $\mathcal I_v$ the remaining issues, and $\mathcal T_v^\star$ the
reconciled subtree passed upward.

Closing an inherited issue requires \emph{concrete recovery evidence}. In the CSV
example, early inspection reports a missing \texttt{summary.json}; a later
subtask writes and reads it back. The JSON parent closes that issue, while the
root retains the unfinished \texttt{chart.png} requirement. Recovery, subtask
completion, and overall completion are judged at their respective scopes.
Language models make semantic judgments; deterministic checks constrain source
coverage, evidence references, issue updates, and key-action preservation.

\par\vspace{-2pt}
\subsection{Rendering Unfinished Requirements}
\vspace{-3pt}

Feedback combines a root overview with local expansion. Completed branches
receive short summaries; incomplete, failed, or unsupported branches expose
shortcuts, issues, and observations. The CSV example summarizes the JSON result
and retains its key write operation, while expanding the missing PNG and failed
plotting evidence. A fresh execution can recreate the JSON and focus further
exploration on the PNG.

Selected critical write operations and final-response information retain their
original content. Task state determines which processes can be summarized and
which issues need expansion, producing high-density feedback of bounded length
for long-horizon tasks. Length budgets and token accounting appear in
Appendix~\ref{app:exp-details}; Appendix~\ref{app:feedback-structure} shows the
final feedback format. Appendix~\ref{app:dense-algorithm} specifies the complete
algorithm, module interfaces, and validation rules.

\par\vspace{-3pt}
\section{REFIT: Evaluating Trajectory Feedback}
\vspace{-2pt}
\label{sec:refit}

REFIT evaluates trajectory feedback by comparing task success on new attempts
with the initial execution. It fixes the agent, tasks, and execution budget;
methods extract feedback from the same raw trajectory, and each method's
subsequent results are compared against this common starting point.

\par\vspace{-2pt}
\subsection{Paired Reruns}
\vspace{-3pt}

\begin{figure}[t]
  \vspace{-4pt}
  \centering
  \includegraphics[trim=8bp 24bp 8bp 6bp,clip,width=0.88\linewidth]{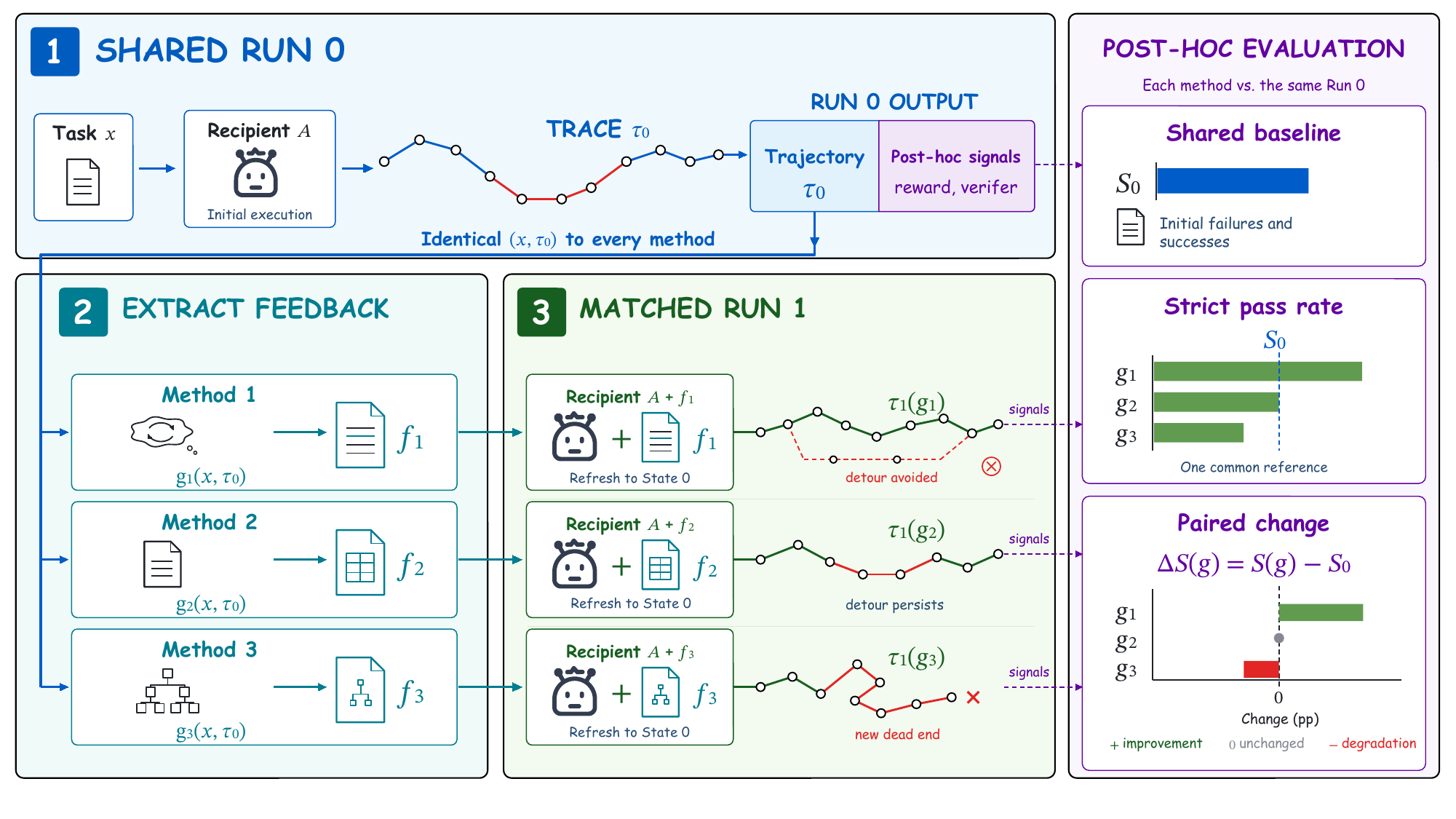}
  \caption{REFIT uses a shared run0 to construct feedback for each method's
  separate run1, starting from initial environment state $s_0$ and a fresh model
  context. Performance changes are
  measured against this common initial execution. Feedback generation is separated
  from post-hoc outcome evaluation.}
  \label{fig:refit-protocol}
  \vspace{-5pt}
\end{figure}

Let $\mathcal D$ be a task distribution, $x\sim\mathcal D$ a task, $A$ the agent
that receives and uses feedback, and $y(x,\tau)\in\{0,1\}$ a strict pass
indicator. An initial attempt (run0) produces a raw trajectory $\tau_0$. For a
fair comparison, all trajectory-feedback methods analyze this same $\tau_0$.
Each generator $g$ produces feedback $f_g$ for its own subsequent attempt (run1):
\begin{equation}
  \tau_0=A(x;\emptyset,s_0),\qquad
  f_g=g(x,\tau_0),\qquad
  \tau_1^g=A(x;f_g,s_0),\qquad
  y_1^g=y(x,\tau_1^g).
  \label{eq:refit-intervention}
\end{equation}
Each execution independently recreates the initial environment state $s_0$ from
the task specification; run1 also uses a fresh model context. Feedback is the
only channel for transferring information between executions.
Figure~\ref{fig:refit-protocol} summarizes the procedure.

\par\vspace{-2pt}
\subsection{Available Information}
\vspace{-3pt}

Feedback generators receive task $x$ and the messages, actions, and observations
in source trajectory $\tau_0$. Under \emph{post-hoc outcome blindness}, they cannot
access post-hoc rewards, hidden tests or verifier outputs, reference answers or
trajectories, or human correctness labels. Verifier outcomes serve only outcome
evaluation, except in the explicitly privileged Verifier Feedback reference.

\par\vspace{-2pt}
\subsection{Measuring Feedback Effectiveness}
\vspace{-3pt}

We measure feedback effectiveness by the change in strict pass probability from
the initial execution. For a fixed agent $A$ and feedback generator $g$, we
define this change as \emph{feedback utility}:
\begin{equation}
  U(g\mid A,\mathcal D)
  =\mathbb E_{x\sim\mathcal D}
  \left[y(x,\tau_1^g)-y(x,\tau_0)\right].
  \label{eq:refit-utility}
\end{equation}
Comparisons fix the recipient, tasks, prompt, tools, and execution budget.
Shared initial executions give gains and subsequent pass rates the same ranking.
Run0 provides a common reference for measuring these changes for a given agent.

\par\vspace{-3pt}
\section{Experiments}
\vspace{-2pt}
\label{sec:exp}

We evaluate task performance, token use, and subsequent behavior on
Terminal-Bench 2.1. The comparisons test complete feedback methods and two core
DENSE designs; an exploratory extension tests repeated feedback, and the case
analysis examines how feedback relates to the recipient's actions.

\par\vspace{-2pt}
\subsection{Experimental Setup}
\vspace{-3pt}

\paragraph{Tasks.}
We use Terminal-Bench 2.1 \citep{merrill2026terminalbench} to study trajectory
reuse in complex, long-horizon tasks. Its containerized terminal environments
require sustained tool interaction across multiple steps, approximating practical
agent workflows. The 89 tasks span software engineering, file and data processing,
and scientific computing, providing varied skills and failure modes within one
benchmark.

\par\vspace{-3pt}
\paragraph{Recipient models.}
We evaluate MiniMax-M2.7 \citep{chen2026minimaxm2},
DeepSeek V4 Pro \citep{deepseek2026v4}, GPT-5.5 \citep{openai2026gpt55},
and Kimi K2.6 \citep{moonshot2026kimik26} as agents receiving and using feedback
(\emph{recipients}). These include open-weight and closed models to compare
their use of trajectory feedback. MiniMax, GPT, and Kimi
use all 89 tasks; DeepSeek uses the 81 supported by its text endpoint, excluding
eight tasks requiring direct image or video input.

\par\vspace{-3pt}
\paragraph{Execution settings.}
We follow REFIT (Section~\ref{sec:refit}) with three paired repetitions per task
and recipient. Within each repetition, trajectory-feedback methods share one
fixed run0 and provide feedback for separate run1 executions. For each recipient,
run0 and run1 use the same task instruction, tools, and execution budget.

\par\vspace{-3pt}
\paragraph{Feedback generation model.}
All-at-once, Step-by-step, DENSE, and Advisor use Gemini 3 Flash Preview to
hold the feedback model fixed while comparing information inputs and extraction
procedures. All conditions use the same complete-feedback length limit;
configurations and prompts appear in Appendices~\ref{app:exp-details}
and~\ref{app:prompts}.

\par\vspace{-3pt}
\paragraph{Metrics.}
Strict pass requires verifier reward exactly 1. Let $y_0$ and $y_1^g$ be the
strict pass indicators for initial and subsequent executions. We define
\begin{equation}
  P_0=\overline{y_0},\qquad P(g)=\overline{y_1^g},\qquad
  \Delta P(g)=P(g)-P_0.
  \label{eq:refit-outcome-change}
\end{equation}
For the single-step comparisons, averages first combine the three repetitions
within each task, then weight tasks equally. Rates are percentages; absolute changes from baseline are in
percentage points (pp), computed as the difference between the two rates.

\par\vspace{-2pt}
\subsection{Comparison Methods}
\vspace{-3pt}

All-at-once analyzes the complete trajectory globally; Step-by-step scores
actions from decision-time prefixes. Advisor tests the feedback model's task
priors; Self-reflection leaves analysis of the raw trajectory to the next-round agent. DENSE
(ours) extracts nested shortcut trees. Verifier Feedback (VF) adds post-hoc signals
as a privileged reference. Table~\ref{tab:feedback-conditions} describes each
method; run0 is the initial-execution reference.

\begin{table}[H]
  \vspace{-4pt}
  \centering
  \caption{Feedback supplied to the subsequent execution. Only Verifier Feedback (VF)
  uses post-hoc reward or verifier output. Advisor denotes the instruction-only
  advisor.}
  \label{tab:feedback-conditions}
  \small
  \setlength{\tabcolsep}{4pt}
  \renewcommand{\arraystretch}{1.08}
  \begin{tabular}{L{.22\linewidth}L{.54\linewidth}L{.15\linewidth}}
    \toprule
    Method & Feedback and construction & Role \\
    \midrule
    Advisor & Advice based only on the task description, without the execution
    trajectory. & Task-prior control \\
    Self-reflection & Raw trajectory for the next-round agent to summarize past
    experience and reflect on it. & Self-reflection baseline \\
    All-at-once & One call analyzes the full trajectory and returns action-level
    assessments and rationales. & Global baseline \\
    Step-by-step & One call per action assesses history through that action,
    excluding its result and later events. & Sequential baseline \\
    \textbf{DENSE (ours)} & Nested subtasks compress redundant attempts while retaining key
    evidence and unresolved issues. & Proposed method \\
    \midrule
    Verifier Feedback & The initial trajectory, its final reward, and verifier
    test output. & Privileged ref. \\
    \bottomrule
  \end{tabular}
  \vspace{-5pt}
\end{table}

All-at-once and DENSE analyze complete-trajectory action outcomes; Step-by-step
scoring excludes the target action's result and later events. Delivered feedback
combines source information with varied analyses, covering self-reflection,
action assessment, and hierarchical extraction.

\par\vspace{-2pt}
\subsection{Task Performance}
\vspace{-3pt}

Table~\ref{tab:full-outcomes} compares strict pass rates. All-at-once and
Step-by-step outperform Advisor on all four recipients and Self-reflection on MiniMax,
GPT, and Kimi. All-at-once ranks second among non-privileged feedback methods
on MiniMax and Kimi; the two methods tie for second on GPT. On DeepSeek,
Step-by-step reaches 65.43\%, above Self-reflection's 63.79\%, while All-at-once
reaches 62.14\%. Analyzing actions thus yields higher strict pass rates than
providing the raw trajectory alone in most
comparisons, with outcomes also depending on the extraction procedure.

DENSE achieves the highest strict pass rate among the tested non-privileged
methods on \textbf{all four recipients}: 52.43\%, 73.25\%, 75.66\%, and 50.94\% on
MiniMax, DeepSeek, GPT, and Kimi, respectively. These improve on the common initial
execution by 7.87, 21.81, 7.12, and 11.61 pp.

\begin{table}[H]
  \vspace{-4pt}
  \centering
  \caption{Strict pass rate (\%) / change from baseline (pp),
  computed before rounding. Baseline aggregates the initial executions shared
  by all methods.
  Verifier Feedback (VF)\textsuperscript{$\dagger$} has privileged outcome information.
  Bold / underline mark the highest / second-highest rates per recipient,
  including Baseline and excluding VF; ties share a mark.}
  \label{tab:full-outcomes}
  \footnotesize
  \setlength{\tabcolsep}{4.5pt}
  \renewcommand{\arraystretch}{1.10}
  \begin{tabular}{@{}lcccc@{}}
\toprule
\rowcolor{refitheader}
Method & MiniMax-M2.7 & DeepSeek V4 Pro & GPT-5.5 & Kimi K2.6 \\
\midrule
\rowcolor{refitmodel}
Baseline & \underline{44.57} & 51.44 & 68.54 & 39.33 \\
Advisor & 41.20 / -3.37 & 53.91 / +2.47 & 69.29 / +0.75 & 43.82 / +4.49 \\
Self-reflection & 38.58 / -5.99 & 63.79 / +12.35 & 72.28 / +3.75 & 46.07 / +6.74 \\
All-at-once & \underline{44.57 / +0.00} & 62.14 / +10.70 & \underline{73.03 / +4.49} & \underline{47.94 / +8.61} \\
Step-by-step & 42.70 / -1.87 & \underline{65.43 / +13.99} & \underline{73.03 / +4.49} & 46.82 / +7.49 \\
\rowcolor{refitours}
\textbf{DENSE (ours)} & \textbf{52.43 / +7.87} & \textbf{73.25 / +21.81} & \textbf{75.66 / +7.12} & \textbf{50.94 / +11.61} \\
\midrule
\rowcolor{refitpriv}
Verifier Feedback\textsuperscript{$\dagger$} & 44.94 / +0.37 & 66.67 / +15.23 & 83.90 / +15.36 & 53.93 / +14.61 \\
\bottomrule
\end{tabular}

  \vspace{-5pt}
\end{table}

On MiniMax, Self-reflection and Step-by-step fall below baseline by 5.99 and 1.87
pp, while All-at-once matches it. DENSE is the only
non-privileged feedback method with a positive gain, exceeding even Verifier
Feedback by 7.49 pp. Appendix~\ref{app:minimax-vf} decomposes this gap and
examines how feedback guides further tuning and delivery on \texttt{tune-mjcf}. Verifier Feedback
exceeds DENSE on GPT and Kimi but trails it by 6.58 pp on DeepSeek,
illustrating how the benefit of outcome information varies across agents.

\begin{figure}[!tbp]
  \vspace{-4pt}
  \centering
  \includegraphics[width=\linewidth]{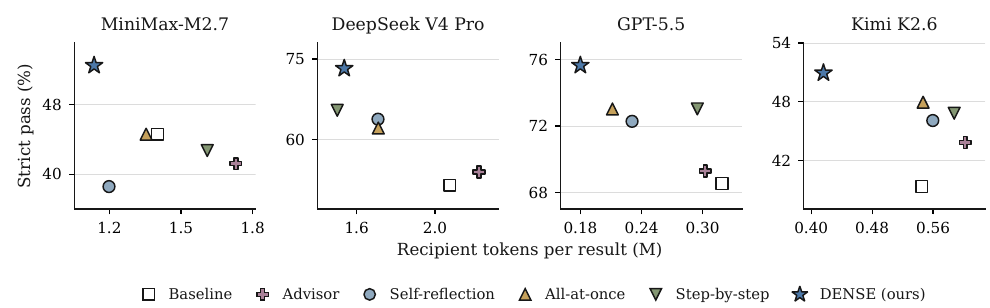}
  \caption{Strict pass rate versus observed recipient tokens per result for non-privileged
  conditions. Means use three repetitions and equal task weights: 89 tasks,
  or 81 for DeepSeek. Baseline is run0; other points count run1 execution only,
  excluding feedback generation and source run0. Axes vary by model; upper left
  is better. Token accounting follows Appendix~\ref{app:exp-details}.}
  \label{fig:performance-tokens}
  \vspace{-5pt}
\end{figure}

\par\vspace{-2pt}
\subsection{Computational Cost}
\vspace{-3pt}

Feedback methods emphasize different reusable experience from the same run0,
guiding exploration in run1. To compare resource use across differently priced
models, we measure tokens: execution input, output, and cache reads and writes.
Appendix~\ref{app:exp-details} details aggregation and attribution.

DENSE reduces observed recipient tokens relative to the displayed baseline
by 19.0\%, 25.9\%, 43.6\%, and 23.7\% for MiniMax, DeepSeek, GPT,
and Kimi, respectively (Table~\ref{tab:full-tokens}). Together with the pass-rate
gains, these results show that subsequent executions complete more tasks with
fewer observed tokens.

\begin{table}[!tbp]
  \vspace{-4pt}
  \centering
  \caption{Observed recipient token use: mean execution tokens (thousands)
  / percentage change from the baseline row. Negative changes indicate reduced
  use.
  \textsuperscript{$\dagger$}Privileged reference.
  Bold / underline mark the lowest / second-lowest token use per recipient,
  excluding VF; ties share a mark.}
  \label{tab:full-tokens}
  \footnotesize
  \setlength{\tabcolsep}{4.5pt}
  \renewcommand{\arraystretch}{1.10}
  \begin{tabular}{@{}lcccc@{}}
\toprule
\rowcolor{refitheader}
Method & MiniMax-M2.7 & DeepSeek V4 Pro & GPT-5.5 & Kimi K2.6 \\
\midrule
\rowcolor{refitmodel}
Baseline & 1,402.2 & 2,075.0 & 319.1 & \underline{545.3} \\
Advisor & 1,729.6 / +23.4\% & 2,224.4 / +7.2\% & 302.6 / -5.2\% & 602.2 / +10.4\% \\
Self-reflection & \underline{1,198.3 / -14.5\%} & 1,709.0 / -17.6\% & 230.7 / -27.7\% & 559.7 / +2.6\% \\
All-at-once & 1,353.9 / -3.4\% & 1,710.5 / -17.6\% & \underline{211.4 / -33.7\%} & 546.9 / +0.3\% \\
Step-by-step & 1,609.8 / +14.8\% & \textbf{1,500.9 / -27.7\%} & 294.8 / -7.6\% & 587.8 / +7.8\% \\
\rowcolor{refitours}
\textbf{DENSE (ours)} & \textbf{1,136.0 / -19.0\%} & \underline{1,536.8 / -25.9\%} & \textbf{179.9 / -43.6\%} & \textbf{416.1 / -23.7\%} \\
\midrule
\rowcolor{refitpriv}
Verifier Feedback\textsuperscript{$\dagger$} & 1,718.5 / +22.6\% & 1,835.0 / -11.6\% & 260.2 / -18.5\% & 682.0 / +25.1\% \\
\bottomrule
\end{tabular}

  \vspace{-5pt}
\end{table}

\Needspace{4\baselineskip}
Figure~\ref{fig:performance-tokens} jointly compares task performance and
recipient token use. DENSE combines the highest non-privileged pass rate with
the lowest execution-token use on MiniMax, GPT, and Kimi. On DeepSeek,
Step-by-step uses fewer execution tokens, while DENSE achieves a higher pass rate.

\newcommand{\denseAblationN}{267}
\newcommand{\denseAblationSensitivityN}{264}
\newcommand{\denseAblationBaselineTokensK}{319.1}
\newcommand{\denseAblationNoShortcutTokensK}{228.6}
\newcommand{\denseAblationSingleRootTokensK}{199.2}
\newcommand{\denseAblationFullTokensK}{179.9}
\newcommand{\denseAblationBaselinePasses}{183}
\newcommand{\denseAblationBaselineRate}{68.54}
\newcommand{\denseAblationBaselineReward}{0.7901}
\newcommand{\denseAblationBaselineSensitivityRate}{69.32}
\newcommand{\denseAblationNoShortcutPasses}{193}
\newcommand{\denseAblationNoShortcutRate}{72.28}
\newcommand{\denseAblationNoShortcutReward}{0.8208}
\newcommand{\denseAblationNoShortcutSensitivityRate}{72.73}
\newcommand{\denseAblationSingleRootPasses}{194}
\newcommand{\denseAblationSingleRootRate}{72.66}
\newcommand{\denseAblationSingleRootReward}{0.8245}
\newcommand{\denseAblationSingleRootSensitivityRate}{73.48}
\newcommand{\denseAblationFullPasses}{202}
\newcommand{\denseAblationFullRate}{75.66}
\newcommand{\denseAblationFullReward}{0.8449}
\newcommand{\denseAblationFullSensitivityRate}{76.14}
\newcommand{\denseAblationBaselineGain}{7.12}
\newcommand{\denseAblationNoShortcutGap}{3.37}
\newcommand{\denseAblationNoShortcutCILow}{-1.50}
\newcommand{\denseAblationNoShortcutCIHigh}{8.24}
\newcommand{\denseAblationNoShortcutSensitivityGap}{3.41}
\newcommand{\denseAblationSingleRootGap}{3.00}
\newcommand{\denseAblationSingleRootCILow}{-1.87}
\newcommand{\denseAblationSingleRootCIHigh}{8.24}
\newcommand{\denseAblationSingleRootSensitivityGap}{2.65}

\par\addvspace{4pt}\noindent\begin{minipage}{\linewidth}
\setlength{\parskip}{2pt}
\par\vspace{-2pt}
\subsection{Ablation Study}
\label{sec:ablation}
\vspace{-3pt}

\begin{wraptable}{r}{0.48\textwidth}
  \vspace{-\intextsep}
  \centering
  \caption{GPT-5.5 ablations: mean execution tokens (K) and strict pass (\%).}
  \label{tab:dense-ablation}
  \small
  \setlength{\tabcolsep}{8pt}
  \renewcommand{\arraystretch}{1.12}
  \begin{tabular}{@{}lrr@{}}
    \toprule
    \rowcolor{refitheader}
    Method & \shortstack{Token cost\\(K)$\downarrow$} & \shortstack{Strict pass\\(\%)$\uparrow$} \\
    \midrule
    run0 & \denseAblationBaselineTokensK & \denseAblationBaselineRate \\
    w/o S\&R & \denseAblationNoShortcutTokensK & \denseAblationNoShortcutRate \\
    w/o hierarchy & \denseAblationSingleRootTokensK & \denseAblationSingleRootRate \\
    \rowcolor{refitours}
    \textbf{DENSE} & \textbf{\denseAblationFullTokensK} & \textbf{\denseAblationFullRate} \\
    \bottomrule
  \end{tabular}
  \vspace{-4pt}
\end{wraptable}

Using GPT-5.5 on all 89 tasks with three paired repetitions, we compare two
ablations under the same REFIT settings. Without S\&R, we retain the nested
subtask tree but omit shortcut construction and issue reconciliation. Without
hierarchy, we replace nested analysis with a single-root, whole-trajectory summary.

The ablations reduce strict pass by \denseAblationNoShortcutGap{} and
\denseAblationSingleRootGap{} pp, respectively, relative to full DENSE, which
also uses fewer observed recipient tokens (Table~\ref{tab:dense-ablation}).
These results support combining both designs.
Appendix~\ref{app:dense-ablation-details} provides implementation details
and a verifier sensitivity check.
\par\WFclear
\end{minipage}\par

\Needspace{4\baselineskip}
\par\vspace{-2pt}
\subsection{Iterative Agent Refinement}
\label{sec:iterative-refinement}
\vspace{-3pt}

\begin{wrapfigure}{R}{0.48\textwidth}
  \vspace{-\intextsep}
  \centering
  \includegraphics[width=\linewidth]{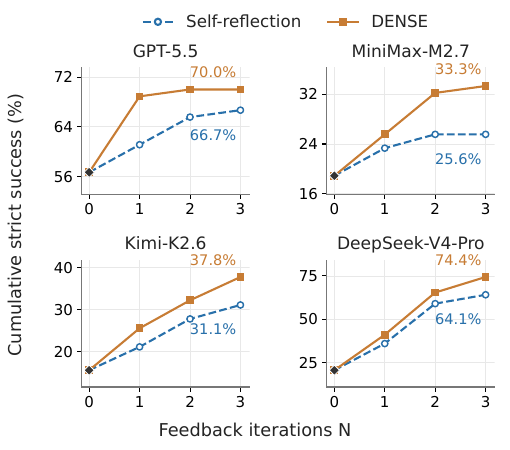}
  \caption{Cumulative strict success on Terminal-Bench 2.1 hard tasks through three feedback rounds.}
  \label{fig:iterative-hard}
  \vspace{-4pt}
\end{wrapfigure}

An exploratory extension compares DENSE with Self-reflection under repeated
execution on the hard subset of Terminal-Bench 2.1,\footnote{Tasks labeled \texttt{hard}
in the official Terminal-Bench 2.1 difficulty annotations.}
using the same four recipients. Appendix~\ref{app:iterative-refinement} details
the execution protocol and paired coverage.

Following the cumulative solved-task reporting in Reflexion
\citep{shinn2023reflexion}, we report \emph{cumulative pass rate}: the fraction
of task--repetition pairs solved at least once within the allowed attempts.

After three feedback iterations, DENSE exceeds Self-reflection by 3.3, 7.8, 6.7,
and 10.3 pp on GPT-5.5,
MiniMax-M2.7, Kimi K2.6, and DeepSeek V4 Pro, respectively
(Figure~\ref{fig:iterative-hard}). These results suggest that structured trajectory
distillation can support iterative agent refinement beyond a single feedback step.

\par\WFclear

\par\vspace{-3pt}
\section{Discussion}
\vspace{-2pt}
\label{sec:discussion}

Recipient differences, feedback comparisons, and an illustrative case clarify
what experience online trajectories contain and how DENSE makes it useful.

\par\vspace{-2pt}
\subsection{Differences in Recipients' Use of Feedback}
\vspace{-3pt}

DENSE's gain over run0 ranges from 7.12 pp on GPT to 21.81 pp on DeepSeek.
MiniMax gains 7.87 pp and exceeds Verifier Feedback (VF) by 7.49 pp.
VF adds post-hoc reward and verifier output to raw history. The paired analysis
in Appendix~\ref{app:minimax-vf} shows that DENSE's advantage mainly reflects
more retained successes in fresh executions.

On \texttt{tune-mjcf}, DENSE distinguishes prior tuning progress from the remaining
speed target while retaining the physics-accuracy requirement. The recipient
tests configurations, writes a tuned model, and verifies it in 25 tool calls,
reaching approximately 55\% of the reference simulation time. The VF recipient
also makes 22 tool calls but continues exploring parameters without producing
the required model file. This contrast illustrates how trajectory feedback can
connect prior exploration to remaining goals and carry execution through to a
verified deliverable.

\par\vspace{-2pt}
\subsection{Useful Experience in Unsupervised Online Trajectories}
\vspace{-3pt}

All-at-once, Step-by-step, and DENSE outperform task-prior-only Advisor on all
four recipients, showing that analyzed traces are more useful than general
advice here. The first two also improve on Self-reflection in most
comparisons; DENSE leads it on MiniMax, DeepSeek, GPT, and Kimi by 13.86, 9.47,
3.37, and 4.87 pp, respectively. Thus, trajectories without post-hoc labels
still contain useful experience, and selecting, connecting, and organizing it
can help recipients use it more effectively than self-reflection on raw trajectories alone. Considerable
room remains to improve how trajectory evidence is selected, connected, and
organized, and how its presentation is adapted to different recipients.

The iterative extension (Section~\ref{sec:iterative-refinement}) suggests that
structured trajectory distillation could provide useful feedback for broader
recursive self-improvement (RSI) systems. Here, the recipient and distillation
procedure remain fixed; improving the feedback mechanism itself and transferring
gains across tasks remain open questions.

\par\vspace{-2pt}
\subsection{From Local Discoveries to Task Completion}
\vspace{-3pt}
\label{sec:feedback-behavior}

The \texttt{password-recovery} task requires recovering a password from a data
image and writing it to a specified file. The initial trajectory contains two
password fragments, but execution continues exploring without writing the file.
DENSE separates failed reconnaissance from successful fragment discovery into
subtasks, retaining failed approaches and reusable evidence in shortcuts. The
parent connects fragment concatenation to the unfinished file-writing requirement,
turning local discoveries into a concrete delivery step.

In a matched comparison with Kimi K2.6, the DENSE-guided execution checks the
environment, verifies the fragments, and writes the result in \textbf{three tool calls},
passing verification. The other four feedback conditions use \textbf{25--47 calls} and
leave the file absent; All-at-once also assembles the same candidate but does not
write it. This contrast illustrates the value of connecting available discoveries
to unfinished task requirements so that reusable evidence supports concrete
completion steps. Appendix~\ref{app:feedback-behavior} provides full feedback,
action records, and outcomes across repetitions.

\par\vspace{-3pt}
\section{Related Work}
\vspace{-2pt}
\label{sec:related}

\par\vspace{-2pt}
\subsection{Trajectory Reuse and Refinement}
\vspace{-3pt}

Prior work uses trajectories to guide later attempts through reflection and
debugging \citep{shinn2023reflexion,zhu2025agentdebug,zhu2026agentdebugx,ma2025dover},
retrieve experience for future tasks
\citep{fang2026trajectorymemory,zhao2024expel,liu2024pract}, and reduce execution
context \citep{ren2026taco,xiao2026agentdiet}. Reward-verified retries also
support training \citep{shi2026r3l}, while successful rollouts inform planning
\citep{gonzalezpumariega2026reliability}.
DENSE extracts reusable evidence without post-hoc supervision by organizing
trajectories hierarchically and summarizing subtask states.

\par\vspace{-2pt}
\subsection{Process Evaluation and Feedback Effectiveness}
\vspace{-3pt}

Process evaluation assesses execution quality
\citep{fan2026agentprocessbench,lu2025agentrewardbench,li2026toolprmbench,zhai2026guide}
and attributes failures to agents or steps
\citep{zhang2025whowhen,liu2026whowhenpro,barke2026agentrx,chen2026trajectoryattribution}.
Critique evaluation measures whether feedback improves subsequent answers
\citep{tang2025realcritic,yu2025rco,sui2026conl}, a criterion that can diverge
from explanation ratings \citep{kunz2022downstream}. Recovery studies examine
restarts with carried-over changes or reconstructed failure states
\citep{wang2026failfastrestartsmart,tan2025recoverybench}; self-correction studies
highlight the need to control resampling and privileged information
\citep{kamoi2024selfcorrection,iscan2026falsification}.
REFIT compares feedback methods using a shared initial trajectory and matched
rerun conditions, measuring changes in strict pass rate without exposing
post-execution outcomes to feedback generation.

\par\vspace{-3pt}
\section{Conclusion}
\vspace{-2pt}
\label{sec:conclusion}

We study how execution traces guide new attempts at the same task without
external outcome labels. DENSE organizes useful steps, evidence of resolved
errors, and unfinished requirements in nested shortcut trees. REFIT compares
task success rates against initial executions, using shared source trajectories
and reset environments and model contexts.
On Terminal-Bench 2.1, DENSE achieves the highest strict
pass rate among tested non-privileged methods across four agent models, improving
over initial executions by 7.12--21.81 pp with 19.0--43.6\% fewer observed
agent tokens in new attempts. GPT-5.5 ablations support combining nested subtask
analysis with shortcut construction and issue reconciliation. On MiniMax,
organized evidence outperforms raw history augmented with verifier signals.
An exploratory hard-task extension shows higher cumulative
pass rates than Self-reflection after three feedback iterations across all four models.
Our work points to promising research directions in the analysis of unlabeled
online trajectories, with potential implications for agent recursive self-improvement.

\par\vspace{-3pt}
\section*{Limitations}
\vspace{-2pt}

Our evaluation is limited to one benchmark, Terminal-Bench 2.1. Nevertheless,
its diverse tasks span software engineering, data processing, and scientific
computing while approximating practical agent workflows. Gains across four
recipients demonstrate DENSE's effectiveness in this varied setting and support
our central insight: trajectories without post-hoc supervision can yield useful
experience. Future work can validate these findings across more benchmarks and
environments.

DENSE's multi-stage distillation adds computation, and feedback may increase
rerun context. A lower-priced generator can offset this overhead: with
Gemini 3 Flash Preview and GPT-5.5, DENSE feedback costs about \$0.14, while
feedback plus rerun costs 36.6\% less than run0 and strict pass rate is
7.12 pp higher (Appendix~\ref{app:feedback-cost}). This tradeoff depends
on model prices and execution costs; reducing generation overhead remains a
direction for improvement.

\par\refitRestoreSpacing

\clearpage
\subsection*{AI use statement}

Generative AI tools assisted with drafting sections of the manuscript,
revising and polishing the text, and retrieving and discovering relevant
literature and research materials. They also assisted with organizing case-study
evidence, preparing tables and figures, LaTeX formatting, code implementation,
and quality review of code and research artifacts. The authors reviewed all
final code, experimental procedures, and resulting data
and artifacts, personally ran the experiments, and collected and organized the
results. The authors checked the case-study descriptions against the underlying
execution traces and verified the reported numerical results against the
experimental records. All AI-assisted content was reviewed by the authors,
who take full responsibility for the final text, claims, code, and artifacts.

\subsection*{Reproducibility statement}

The anonymous supplementary materials provide code, prompts, configurations,
and data for reproducing the reported experiments. Sections~\ref{sec:refit}
and~\ref{sec:exp} describe the evaluation protocol;
Appendices~\ref{app:exp-details}, \ref{app:prompts}, and~\ref{app:dense-algorithm}
provide experimental settings, metric definitions, prompts, and algorithm details.

\subsection*{Ethics statement}

This study evaluates agents in the containerized environments of
Terminal-Bench 2.1 and does not involve human-subject experiments. Execution
traces may contain sensitive information in real-world deployments; applying
DENSE to such traces would require appropriate access controls and removal of
sensitive content. Distilled feedback may preserve errors or unsafe actions
from the source trajectory, so improved benchmark performance should not be
interpreted as a guarantee of safe deployment.

\clearpage
\bibliography{references}
\bibliographystyle{iclr2027_conference}

\clearpage
\appendix
\Needspace{6\baselineskip}
\section{Experimental Configuration and Accounting}
\label{app:exp-details}

\subsection{Feedback Configuration and Budget}
\label{app:feedback-configuration}
All generated-feedback conditions use Gemini 3 Flash Preview
(API model identifier: \texttt{gemini-3-flash-preview}).
DENSE builds its initial subtask tree from the full trajectory, scoring actions
before producing subtask summaries. It then compresses consecutive attempts under
the same parent, reconciles issue states from children to parents, and organizes
the final feedback around unresolved issues. Tree feedback is limited to
180~KiB within a total feedback budget of 200~KiB. Other methods use the same
total budget. When shortening feedback, DENSE operates on complete semantic
blocks to avoid cutting through evidence.
Appendix~\ref{app:prompts} presents the prompt roles, input views, and
instructions used to construct and deliver these feedback representations.

\subsection{DENSE Ablation Details}
\label{app:dense-ablation-details}
The ablations in Section~\ref{sec:ablation} use GPT-5.5 with high reasoning effort
and Gemini 3 Flash Preview as the feedback model. The hierarchical control keeps
the full-trajectory tree builder and its score-then-summarize procedure; only
the subsequent shortcut and reconciliation stages are removed. The single-root
control instead makes one logical analysis call over the task and all recorded
action--observation text, requesting completion status, evidence, lessons, and
next steps. Both use only the task and visible trajectory, without post-hoc
reward or verifier output.

The analysis output limit is 8,192 tokens. One single-root analysis
(\texttt{dna-assembly}, repetition 3) required a 16,384-token limit to obtain a
valid response with the same prompt. Retries address unscored analysis or
execution failures; the first valid execution result is retained, including
zero and partial rewards. All four conditions have complete coverage of the
same \denseAblationN{} task--repetition pairs.

Token cost in Table~\ref{tab:dense-ablation} counts recipient execution only,
using the accounting in Appendix~\ref{app:recipient-token-accounting}; feedback
generation is excluded. All 1,068 results have recorded usage; 64 report
completed calls only, omitting unfinished calls. The table therefore reports
observed token use.

Historical verifier logs for \texttt{torch-pipeline-parallelism} contain port
binding conflicts in ten scored runs across the four conditions, including
two DENSE runs. We retain those recorded results in the main table and check
sensitivity by excluding all three repetitions of this task from every
condition. On the remaining \denseAblationSensitivityN{} pairs, strict pass
rates are \denseAblationBaselineSensitivityRate\% for run0,
\denseAblationNoShortcutSensitivityRate\% without shortcut/reconcile,
\denseAblationSingleRootSensitivityRate\% for the single-root summary, and
\denseAblationFullSensitivityRate\% for full DENSE. The DENSE gaps are
\denseAblationNoShortcutSensitivityGap{} and
\denseAblationSingleRootSensitivityGap{} pp, respectively.

\subsection{Recipient Token Accounting}
\label{app:recipient-token-accounting}
Recipient tokens sum uncached input, cache-read input, cache-creation input, and
output. Relative change is
$(\overline{T}_{g}/\overline{T}_{0}-1)$, reported as a percentage, where $\overline{T}_{0}$ is the
mean in the displayed baseline row and $\overline{T}_{g}$ is the method mean.
Both average repetitions within tasks and then give applicable tasks equal weight. Recipient usage is recorded
for all 7,308 outcomes: 5,554 records report execution totals and 1,754 sum
completed calls only. The latter
omit unfinished calls, and unknown usage is not replaced with zero.

\subsection{Feedback Cost Attribution}
\label{app:feedback-attribution}

Combined cost attributes one feedback generation to each subsequent execution
and adds its recipient usage. It excludes the initial execution unless explicitly
stated. Reused feedback is attributed to each result without implying new
generation spending, so these per-result estimates are not total experimental
expenditure.

\Needspace{6\baselineskip}
\subsection{Iterative Refinement Protocol and Coverage}
\label{app:iterative-refinement}

The exploratory analysis in Section~\ref{sec:iterative-refinement} reports
run0 through run3 on the 30 tasks with official difficulty label \texttt{hard}.
These comprise the initial execution and three feedback-conditioned attempts.
DeepSeek uses the 26 supported by its text endpoint. Each task has three
repetitions. Task instructions, tools, and per-attempt execution limits are
matched between methods. DENSE uses Gemini 3 Flash Preview for distillation;
Self-reflection supplies the corresponding raw action--observation history for
the next-round agent to summarize experience and reflect on previous attempts.

Recipient models and feedback procedures remain fixed across iterations.
Every attempt resets the environment and model context. The explicit memory
contains feedback from up to the three preceding attempts ($K=3$); each new item uses
only the immediately preceding execution. Previously injected feedback is
removed from the trajectory view used to construct new feedback, avoiding
recursive inclusion of older memory. Post-hoc rewards and verifier outputs are
excluded from feedback and recipient inputs. Cumulative outcomes are computed
retrospectively; no verifier-based early stopping is required to obtain them.

For recipient $m$ and method $g$, let $P_m$ be a fixed set of matched
task--repetition pairs and $y^{g}_{m,t,r,n}$ indicate strict success at iteration
$n$. We compute
\begin{equation}
  C_{m,g}(N)=\frac{1}{|P_m|}\sum_{(t,r)\in P_m}
  \mathbf{1}\!\left[\max_{0\leq n\leq N}y^{g}_{m,t,r,n}=1\right],
  \qquad N\in\{0,1,2,3\}.
  \label{eq:iterative-cumulative-success}
\end{equation}
Each repetition is evaluated separately; successes are not pooled across
repetitions.

All supported hard-task pairs are included: 90 for GPT, MiniMax, and Kimi,
and 78 for DeepSeek. All executions through run3 are complete for both
methods. The figure weights task--repetition pairs equally; with all three
repetitions available per task, this is equivalent to averaging repetitions
within each task and then weighting tasks equally, as in the main comparisons.

These are descriptive comparisons within a three-round refinement horizon.
Equal attempt limits do not imply equal token or monetary costs. Cumulative
success measures whether a task has been solved within that horizon, not the
accuracy of the latest attempt. Without an independent-retry control, the
increase over run0 cannot be attributed entirely to feedback; the comparison
here concerns DENSE relative to Self-reflection.

\Needspace{6\baselineskip}
\section{Supplementary Cost Analysis}
\label{app:supplementary-results}

We report a priced cost breakdown for GPT-5.5. Aggregation and feedback attribution follow
Appendix~\ref{app:exp-details}.

\subsection{Estimated Feedback and Rerun Costs for GPT-5.5}
\label{app:feedback-cost}
\newcommand{\gptBaselineCost}{1.81}
\newcommand{\gptDenseRerunCost}{1.01}
\newcommand{\gptDenseFeedbackCost}{0.14}
\newcommand{\gptDenseTotalCost}{1.15}
\newcommand{\gptDenseWorkflowCost}{2.96}
\newcommand{\gptDenseFeedbackBaselinePct}{7.7}
\newcommand{\gptDenseFeedbackRerunPct}{13.9}
\newcommand{\gptDenseTotalSavingPct}{36.6}

We estimate GPT-5.5 execution and Gemini 3 Flash Preview feedback costs using
standard base text rates. Input/output prices
per million tokens are \$5/\$30 for GPT-5.5 \citep{openai2026gpt55pricing}
and \$0.50/\$3 for Gemini \citep{google2026geminipricing}. Input and output
usage are priced separately and added. These standardized estimates use
uncached rates, without long-context surcharges or unrecorded usage; actual
charges may differ.

\begin{table}[H]
  \centering
  \caption{Mean estimated USD costs over 89 tasks with three repetitions each.
  Baseline is the initial GPT-5.5 execution; other rows show the GPT-5.5 rerun,
  Gemini feedback, and their sum. $\Delta$ is the sum's percentage change from
  baseline, computed before rounding. Dashes mean no feedback generation.
  Bold / underline mark the best / second-best values: lower cost and higher
  pass rate are better; feedback is ranked among generators.}
  \label{tab:gpt55-priced-costs}
  \footnotesize
  \setlength{\tabcolsep}{5pt}
  \renewcommand{\arraystretch}{1.10}
  \begin{tabular}{@{}lcccc@{}}
\toprule
\rowcolor{refitheader}
Method & Execution (\$) & Generator (\$) & Total (\$) / $\Delta$ & Pass (\%) \\
\midrule
\rowcolor{refitmodel}
Baseline & 1.8085 & --- & 1.8085 & 68.54 \\
Advisor & 1.7067 & \textbf{0.0024} & 1.7091 / -5.5\% & 69.29 \\
Self-reflection & 1.2596 & --- & 1.2596 / -30.4\% & 72.28 \\
All-at-once & \underline{1.1565} & \underline{0.0138} & \underline{1.1704 / -35.3\%} & \underline{73.03} \\
Step-by-step & 1.5947 & 0.1207 & 1.7154 / -5.2\% & \underline{73.03} \\
\rowcolor{refitours}
\textbf{DENSE (ours)} & \textbf{1.0075} & 0.1396 & \textbf{1.1470 / -36.6\%} & \textbf{75.66} \\
\bottomrule
\end{tabular}

\end{table}

DENSE feedback costs \$\gptDenseFeedbackCost{}, or
\gptDenseFeedbackBaselinePct\% of the \$\gptBaselineCost{} baseline and
\gptDenseFeedbackRerunPct\% of its \$\gptDenseRerunCost{} rerun. Feedback plus
rerun costs \$\gptDenseTotalCost{}, \textbf{\gptDenseTotalSavingPct\% below baseline},
while strict pass rate increases from 68.54\% to 75.66\%. In this configuration,
a small additional feedback cost accompanies lower rerun spending and higher
task success. Even with feedback generation included, the estimated cost of this
rerun remains below the initial-execution reference.
Figure~\ref{fig:absolute-token-cost} visualizes this decomposition with an
explicit run0 baseline; its gray segments represent the feedback generator.

Including the initial execution gives \$\gptDenseWorkflowCost{} for the complete
DENSE workflow; cost attribution follows Appendix~\ref{app:feedback-attribution}.

\raggedbottom
\Needspace{6\baselineskip}
\section{Anatomy of DENSE Feedback}
\label{app:feedback-structure}

\subsection{Feedback Interface and Reading Guide}
\label{app:feedback-interface}

DENSE presents an execution as a hierarchy of task progress, reusable evidence,
and remaining problems. Whereas Section~\ref{sec:method} describes its
construction, this appendix examines the artifact delivered to the recipient.
The feedback begins with the fresh-environment reminder reproduced in
Appendix~\ref{app:prompts}: files and processes mentioned in the report belong to
the previous attempt, and required outputs must be recreated. The report then
gives a root summary, nested tasks, compressed attempts, and selected
action--observation evidence.

Table~\ref{tab:case-feedback-fields} summarizes how to read these components.
Task completion and shortcut outcome describe different scopes. A shortcut can
succeed at recovering a local path while its parent task remains incomplete.
Moreover, \texttt{complete} is a judgment supported by the historical trajectory,
not a statement that the hidden verifier passed. Numeric action scores used
during construction are omitted from the selected final feedback.

\begin{table}[H]
  \centering
  \caption{Reading the recipient-visible DENSE artifact. Field names follow the
  emitted report; the explanation describes their role rather than adding
  information to the recipient's input.}
  \label{tab:case-feedback-fields}
  \small
  \setlength{\tabcolsep}{4pt}
  \renewcommand{\arraystretch}{1.12}
  \begin{tabular}{L{.31\linewidth}L{.62\linewidth}}
    \toprule
    Visible component & Interpretation \\
    \midrule
    Root \texttt{summary}, \texttt{state}, and \texttt{completion-scope}
      & Overall historical progress and the scope of the completion judgment. \\
    Nested \texttt{task} blocks
      & Subtask boundaries, summaries, and completion states. \\
    \texttt{shortcut}, \texttt{outcome}, and \texttt{source}
      & A compressed group of attempts, its local result, and source nodes. \\
    \texttt{dead-end} / \texttt{working-path}
      & Unproductive attempts and useful procedures or intermediate results. \\
    \texttt{open-issue} / \texttt{evidence}
      & A remaining problem and the historical evidence supporting it. \\
    \texttt{key-action}, \texttt{tool-call}, and \texttt{observation}
      & Selected executable details and their observed consequences. \\
    \bottomrule
  \end{tabular}
\end{table}

\subsection{Hierarchy, Shortcuts, and Unresolved Issues}
\label{app:feedback-worked-example}

The \texttt{largest-eigenval} task requires a dominant eigenpair solver for small
real matrices, including cases with complex eigenpairs, that outperforms a NumPy
reference. We inspect Kimi K2.6, repeat 2. The initial execution established
timing baselines and compared analytical formulas, iterative methods, and library
calls. Some local approaches were useful, but manually reconstructing complex
eigenvectors from LAPACK's real-valued representation left a performance
bottleneck for larger sizes.

\begin{figure}[H]
  \centering
  \resizebox{\linewidth}{!}{\begin{tikzpicture}[
  x=1cm,y=1cm,
  every node/.style={font=\sffamily\linespread{1}\fontsize{8}{10}\selectfont,align=left},
  nodebox/.style={draw=black!45,rounded corners=2pt,inner sep=4pt},
  resolved/.style={nodebox,draw=caseresolved,fill=caseresolved!6},
  unresolved/.style={nodebox,draw=caseopen,fill=caseopen!6},
  evidence/.style={nodebox,draw=caseevidence,fill=caseevidence!5},
  edge/.style={-{Latex[length=1.6mm]},draw=black!60,line width=.55pt}
]
  \node[anchor=west,font=\sffamily\bfseries\fontsize{8}{10}\selectfont] at (.05,.85)
    {(a) Source actions};
  \node[nodebox,fill=black!2,text width=4.4cm] at (2.4,0)
    {\textbf{a1--a2}\\Establish timing baselines\\and mathematical constraints};
  \node[nodebox,fill=black!2,text width=7.0cm] at (9.1,0)
    {\textbf{a3--a29}\\Compare solver paths and retain useful results;\\
     diagnose reconstruction overhead and continue probing.};
  \draw[edge] (4.88,0)--(5.25,0);
  \node[anchor=west,font=\sffamily\bfseries\fontsize{8}{10}\selectfont] at (.05,-1.35)
    {(b) Reconciled tree};
  \node[anchor=west,font=\sffamily\bfseries\fontsize{8}{10}\selectfont] at (7.1,-1.35)
    {(c) Rendered feedback (schematic)};
  \node[unresolved,text width=5.7cm] (root) at (3.15,-2.2)
    {\textbf{n29 / n35: incomplete}\\Root task / benchmark and strategy analysis};
  \node[resolved,text width=2.65cm] (done) at (1.55,-4.05)
    {\textbf{n30: complete}\\Baselines and constraints\\a1--a2};
  \node[unresolved,text width=2.65cm] (open) at (4.8,-4.05)
    {\textbf{n34: incomplete}\\Solver optimization\\a3--a29};
  \draw[edge] (root.south)--++(0,-.25)-|(done.north);
  \draw[edge] (root.south)--++(0,-.25)-|(open.north);
  \node[resolved,text width=2.65cm] (s0) at (1.55,-5.85)
    {\textbf{shortcut\_0}\\succeeded\\Sources: n0, n1};
  \node[unresolved,text width=2.65cm] (s5) at (4.8,-5.85)
    {\textbf{shortcut\_5}\\partial\\Sources: n31, n32, n33};
  \draw[edge] (done)--(s0);
  \draw[edge] (open)--(s5);
  \node[anchor=north west,text width=5.9cm,font=\sffamily\fontsize{7}{9}\selectfont]
    at (.1,-6.8)
    {n29 and n35 are grouped here for space; action leaves and other evidence
     are omitted.};
  \node[resolved,text width=5.2cm,anchor=north west] (fold) at (7.1,-1.9)
    {\textbf{Completed branch: compact}\\
     Task summary and selected evidence \textbf{a2}.\\
     No separate shortcut\_0 block.};
  \node[unresolved,text width=5.2cm,anchor=north west] (expand)
    at ([yshift=-3mm]fold.south west)
    {\textbf{Unresolved branch: expanded}\\
     Task-level open issue:\\
     Python complex-vector reconstruction\\[2pt]
     Partial shortcut:\\
     dead-end / working-path / open-issue\\[2pt]
     Retained actions include \textbf{a28}.};
  \node[evidence,text width=5.2cm,anchor=north west]
    at ([yshift=-3mm]expand.south west)
    {\textbf{Read as historical evidence}\\
     Local completion is not a verifier verdict.};
\end{tikzpicture}}
  \caption{Structure and rendering for \texttt{largest-eigenval}, Kimi K2.6,
  repeat 2. This is a simplified view of the actual artifact, with source IDs
  retained and node descriptions shortened. Colors and explicit state labels
  distinguish completed content from unresolved content. The right panel
  paraphrases the rendered organization; Figure~\ref{fig:case-eigen-excerpt}
  provides the original wording.}
  \label{fig:case-eigen-structure}
\end{figure}
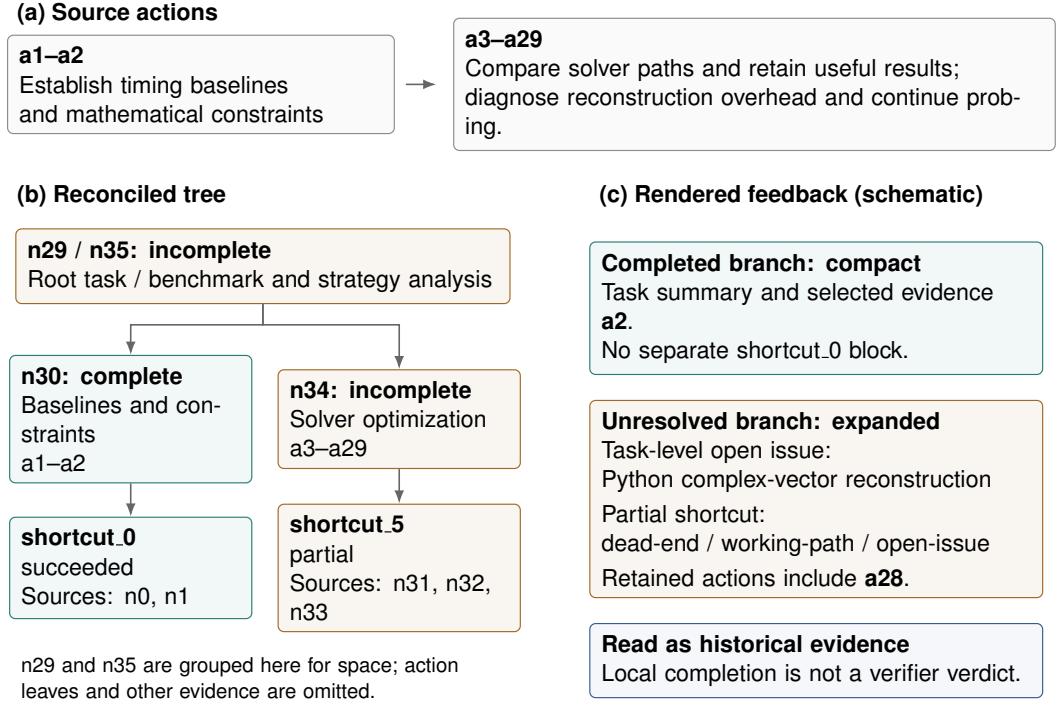

Figure~\ref{fig:case-eigen-structure} relates the source phases, reconciled tree,
and delivered feedback. The baseline-and-constraints branch, n30, is complete;
the solver-optimization branch, n34, remains incomplete. Under n34,
\texttt{shortcut\_5} combines n31--n33 with a \texttt{partial} outcome, retaining
both useful computation paths and the unresolved reconstruction overhead.
The complete branch is rendered compactly: its internal
\texttt{shortcut\_0} is not displayed separately, although the selected baseline
action a2 remains visible. The incomplete branch exposes its issue and supporting
actions, including a28. Thus, the reconciled tree and the final feedback are
related representations with different levels of detail.

The excerpt in Figure~\ref{fig:case-eigen-excerpt} shows how these distinctions
appear in the input itself. An open issue states the bottleneck at the task
level, while the partial shortcut records both unsuccessful approaches and
reusable results. Evidence identifiers connect this synthesis to the retained
actions. These identifiers are references within the source analysis;
a1, a2, and so on enumerate serialized actions, each of which may contain
multiple tool invocations. They are not counts of shell commands or elapsed time.

\begingroup
\par\addvspace{6pt}
  \begin{casebox}{DENSE feedback excerpt: largest-eigenval}
    \caseVerbatim{figures/case_studies/largest_eigenval_excerpt.txt}
  \end{casebox}
  \captionof{figure}{Selected original spans of the delivered feedback. Indentation and
  wrapping are adjusted for readability; \texttt{[...]} explicitly marks omitted
  attributes, blocks, or action contents. The retained wording is unchanged.
  The excerpt is abbreviated markup, not a replacement feedback template.}
  \label{fig:case-eigen-excerpt}
\par\addvspace{6pt}
\endgroup

\subsection{Local Recovery and Remaining Obligations}
\label{app:feedback-nested-state}

The \texttt{gpt2-codegolf} case illustrates why completion must be scoped to a
subtask. In GPT-5.5 repeat 0, the original attempt implemented a compact C
inference program, but generated repeated tokens. One local detour had already
been recovered: unavailable inspection tools were replaced with available
commands. The corresponding \texttt{shortcut\_1} is marked
\texttt{succeeded}, yet its parent n13 is incomplete because the checkpoint
layout is still unverified. A separate shortcut covering implementation and
inference remains \texttt{failed}.

The feedback explicitly preserves the distinction:
\lq\lq{}The specific internal layout and ordering of the GPT-2 weights within the
binary checkpoint file are not yet verified, and the current implementation's
assumptions are failing.''
Successful inspection and compilation therefore remain useful intermediate
results without closing the inference problem. Appendix~\ref{app:feedback-behavior}
examines how the recipient subsequently uses this information.

\Needspace{6\baselineskip}
\section{Feedback and Subsequent Agent Behavior}
\label{app:feedback-behavior}

\subsection{Matched Cases and Information Views}
\label{app:feedback-comparison-setup}

We compare the information supplied by each feedback condition with the
recipient's subsequent tool actions. Each comparison fixes the task, recipient,
repetition, and source run0; feedback is consumed in a fresh run1 under the
protocol in Section~\ref{sec:refit}. We examine Advisor, Self-reflection, All-at-once,
Step-by-step, and DENSE. In these five-condition comparisons, verifier
results are used only to evaluate the completed executions.
Section~\ref{app:minimax-vf} separately examines Verifier Feedback as a
privileged-information condition.

Table~\ref{tab:case-outcomes} reports all three repetitions.

\begin{table}[!htbp]
  \centering
  \caption{Input views and information processing in the five compared conditions.
  The Step-by-step feedback in these cases identifies its view as
  \texttt{causal\_prefix}; the restriction applies to each scoring call.}
  \label{tab:case-information-views}
  \small
  \setlength{\tabcolsep}{4pt}
  \renewcommand{\arraystretch}{1.12}
  \begin{tabular}{L{.16\linewidth}L{.32\linewidth}L{.43\linewidth}}
    \toprule
    Arm & Extractor input and unit & Information delivered to the recipient \\
    \midrule
    Advisor & Task instruction only; task-level advice.
      & General strategy, risks, and a completion checklist. \\
    Self-reflection & Serialized source trajectory; no externally generated analysis.
      & Raw actions and observations for the next-round agent to summarize and reflect on. \\
    All-at-once & One joint analysis of the full trajectory.
      & Action-level assessments and rationales alongside the trajectory. \\
    Step-by-step & One decision-time prefix per target action, excluding its
      result and later events.
      & Action-level assessments with a \lq\lq{}Known at the time'' basis. The
      delivered table also contains the source observations. \\
    DENSE (ours) & Nested subtasks, grouped attempts, and issue reconciliation.
      & Task states, reusable paths, unresolved issues, and selected evidence. \\
    \bottomrule
  \end{tabular}
\end{table}

These views differ in more than the amount of context available to the
extractor. Global access can still produce local action judgments; a prefix
judgment can use outcomes of earlier actions; and a recipient can read the
observations reproduced in a flat feedback table. We therefore distinguish
facts that are present, connections explicitly made by the extractor, and
remaining goals made actionable by the feedback.

\subsection{Shared Evidence, Different Feedback}
\label{app:feedback-information-comparison}

In \texttt{password-recovery}, Kimi K2.6 repeat 0, the task is to recover a
23-character string beginning with \texttt{8XD} and ending with \texttt{W54},
and write a candidate to \texttt{/app/recovered\_passwords.txt}.
The source trajectory already contains two fragments from a data image:
\texttt{8XDP5Q2RT9Z} and \texttt{K7VB3BV4WW54}. Their lengths are 11 and 12;
concatenating them yields a candidate satisfying the stated length and format.
The initial execution nevertheless continued inspecting archive structures and
byte offsets and did not create the required output file.

Table~\ref{tab:case-password-feedback} compares how this shared situation is
expressed. Self-reflection preserves the fragments as observations. Both flat
extractors discuss useful evidence and endorse further inspection at particular
steps. In Step-by-step, the basis for OA table row 23 explicitly names both
fragments; their presence is not unique to DENSE. Advisor also explicitly
requires writing the output, but cannot connect that requirement to evidence
from an attempt it has not observed.

\begin{table}[!htbp]
  \centering
  \caption{Feedback excerpts for the same \texttt{password-recovery}
  run0, with code formatting normalized. OA row numbers refer to the feedback's action--observation table.
  The interpretation column is our analysis, not additional recipient input.}
  \label{tab:case-password-feedback}
  \footnotesize
  \setlength{\tabcolsep}{4pt}
  \renewcommand{\arraystretch}{1.13}
  \begin{tabular}{L{.14\linewidth}L{.47\linewidth}L{.30\linewidth}}
    \toprule
    Arm / source & Retained feedback span & Information expressed \\
    \midrule
    Advisor\newline Strategy 5
      & \lq\lq{}\caseQuoteAdvisor''
      & Explicit delivery requirement, without the discovered fragments. \\
    \addlinespace[3pt]
    Self-reflection\newline Observation 22
      & \texttt{PASSWORD=8XDP5Q2RT9Z}\newline
        \ldots{} \texttt{K7VB3BV4WW54}
      & Both fragments are available; no extractor adds a synthesis. \\
    \addlinespace[3pt]
    All-at-once\newline Reasoning 29
      & \lq\lq{}\caseQuoteGlobal''
      & A positive assessment of locating byte offsets within the search. \\
    \addlinespace[3pt]
    Step-by-step\newline Reasoning 29
      & \lq\lq{}\caseQuotePrefix''
      & Prefix evidence motivates further byte inspection to reconstruct the string. \\
    \addlinespace[3pt]
    DENSE\newline Root issue, n26
      & \lq\lq{}\caseQuoteDense''
      & A concrete assembled candidate is bound to the outstanding deliverable. \\
    \bottomrule
  \end{tabular}
\end{table}

DENSE's root summary explicitly assembles the candidate and explains that the
task remains incomplete because it has not been written out. Its tree separates
unsuccessful reconnaissance from successful fragment discovery while retaining
the delivery obligation above that local success. The salient change is thus a
task-level synthesis: what the available fragments jointly support, and what
still has to be done. A generic inherited search issue also remains in this
artifact; the example does not imply that reconciliation removes every
irrelevant issue. Nor do positive flat action labels assert that the whole task
has succeeded.

\subsection{From Feedback to Subsequent Actions}
\label{app:feedback-action-comparison}

Table~\ref{tab:case-password-actions} records the subsequent execution.
DENSE's recipient lists the fresh environment, reruns \texttt{strings} on the
known image, then writes and reads back the assembled candidate. The run uses
three tool invocations and passes both output-file and password-match checks.
The four comparison runs use 25--47 invocations and leave the required file
absent.

\begin{table}[!htbp]
  \centering
  \caption{Observed behavior in \texttt{password-recovery}, Kimi K2.6, repeat 0.
  Call indices count tool invocations in recorded order; a call can contain
  several shell operations, and parallel calls are counted separately.
  \lq\lq{}File'' means the required output exists at verification.}
  \label{tab:case-password-actions}
  \footnotesize
  \setlength{\tabcolsep}{4pt}
  \renewcommand{\arraystretch}{1.14}
  \begin{tabular}{L{.15\linewidth}L{.56\linewidth}rrr}
    \toprule
    Arm & Recorded action excerpts & Calls & File & Reward \\
    \midrule
    Advisor & Searches filenames (2); inspects image context (15);
      continues to archive-directory signatures (47). & 47 & No & 0 \\
    Self-reflection & Searches for a full contiguous match (3); carves and
      inspects archives; attempts a loopback device (25). & 25 & No & 0 \\
    All-at-once & Scans known image patterns (2); inspects archives and
      offsets; prints the assembled candidate and length 23 (26). & 26 & No & 0 \\
    Step-by-step & Searches full patterns (1--3); inspects fragments,
      extracts archives, and parses the central directory (35). & 35 & No & 0 \\
    DENSE (ours) & Lists the environment (1); rechecks fragments with
      \texttt{strings} (2); writes the candidate and reads it back (3).
      & 3 & Yes & 1 \\
    \bottomrule
  \end{tabular}
\end{table}

The All-at-once trace provides a useful qualification: its final invocation
prints the \emph{same} assembled candidate and confirms its length as 23.
This recipient can infer the answer, but the recorded run ends without the
required file. The contrast concerns the timing of the transition from
investigation to delivery, not an absolute inability of the other feedback
conditions to reconstruct the string. DENSE's explicit remaining obligation
corresponds to that transition occurring immediately after a short check in the
fresh environment. The trace establishes this correspondence; it does not reveal
the recipient's internal attention or isolate the effect of the issue's position.

\paragraph{Carrying forward an unfinished optimization.}
In \texttt{largest-eigenval}, the retained reconstruction bottleneck corresponds
to a subsequent implementation using \texttt{scipy.linalg.eig} with
\texttt{overwrite\_a=True} and \texttt{check\_finite=False}.
The recipient selects the largest-magnitude eigenvalue and its corresponding
vector. This implementation follows from more than the issue summary:
the feedback already retains an a29 experiment with SciPy and
\texttt{overwrite\_a=True}, whose observation records termination with exit code
143. The additional \texttt{check\_finite=False} setting is absent from the
feedback and appears in run1. The recipient therefore completes and extends
an available experimental direction rather than receiving a finished patch.

The flat feedback also contains overhead-related evidence and library
comparisons. In the representative run, DENSE passes 27/27 parameterized checks,
All-at-once 26/27, and Step-by-step 25/27. Only DENSE meets the strict pass
criterion in this repetition. These results support the observed implementation outcome; the
performance checks remain sensitive to timing variation.

\subsection{MiniMax Feedback Use and Outcome Transitions}
\label{app:minimax-vf}

On MiniMax-M2.7, DENSE achieves a strict pass rate of 52.43\%, compared with
44.94\% for Verifier Feedback (VF), a gap of $+7.49$ pp computed before rounding.
We first decompose this difference over all 89 tasks and three repetitions,
then examine how the two feedback conditions guide further optimization and
delivery on \texttt{tune-mjcf}.

\subsubsection{Retaining Success and Repairing Failures}

Of the 267 source executions, 119 strictly succeed and 148 do not. We classify
a subsequent execution as retaining success when both run0 and run1 succeed,
regressing when only run0 succeeds, and repairing when only run1 succeeds.
These are paired outcome descriptions; run1 must recreate the solution under
the fresh-environment protocol, so a regression need not involve editing an
existing artifact.

\begin{table}[!htbp]
  \centering
  \caption{MiniMax outcome transitions relative to the shared source run0.
  Retained and regressed partition the 119 successful sources. Repaired counts
  successes among the 148 unsuccessful sources. Net gain is repaired minus
  regressed. Each method has 267 subsequent executions.}
  \label{tab:minimax-vf-transitions}
  \small
  \setlength{\tabcolsep}{7pt}
  \def\minimaxTransitionHeader{Method & Retained & Regressed & Repaired & Net gain & Successes \\}
  \begin{tabular}{lrrrrr}
\toprule
\minimaxTransitionHeader
\midrule
Self-reflection & 83 & 36 & 20 & $-16$ & 103 \\
\textbf{DENSE (ours)} & 110 & 9 & 30 & $+21$ & 140 \\
VF & 82 & 37 & 38 & $+1$ & 120 \\
\bottomrule
\end{tabular}

\end{table}

VF repairs more unsuccessful sources than DENSE (38 versus 30), but loses
success on many more previously successful sources (37 versus 9).
Consequently, DENSE's advantage is 28 additional retained successes offset by
8 fewer repairs, yielding 20 additional successes overall. Among the 119
initially successful executions, DENSE's regression rate is 7.56\%, compared
with VF's 31.09\%. This lower regression rate shows greater stability when
reusing historical experience to recreate successful outcomes in a fresh
environment.

Actual tool-use records show that 32 of VF's 37 regressions contain no tool
calls; DENSE has no zero-tool execution among its 267 runs. Self-reflection
also exhibits this behavior in 27 of its 36 regressions. Differences also
arise during active execution: among the 49 pairs where DENSE passes and VF
does not, VF uses tools in 17, including 13 with an unsuccessful run0. The
following task illustrates how the recipients differ in turning further
exploration into a verified deliverable.

\subsubsection{Case: Completing a Constrained Simulation Optimization}

\paragraph{Task and shared history.}
The \texttt{tune-mjcf} task asks the agent to tune a MuJoCo model so that
simulating two seconds takes at most 60\% of the reference execution time.
The full final physics state must remain within an absolute tolerance of
$10^{-5}$, without NaN or Inf. The reference file
\texttt{/app/model\_ref.xml} must remain unchanged; the tuned model must be
saved as \texttt{/app/model.xml}. A provided \texttt{eval.py} script supports
local evaluation.

The shared source execution explores solver parameters, timestep changes,
and simulation flags. Its delivered model passes the reference-integrity,
file-existence, and physics-correctness checks, but fails the speed check:
the measured simulation time is approximately 110\% of the reference time.
Both feedback conditions use this history, and their recipients start from
fresh environments under the same execution limits.

\paragraph{Feedback distinguishes partial progress from the remaining target.}
VF supplies the source trajectory, its reward of 0.75, and the verifier
output identifying the unmet speed requirement. DENSE instead organizes the
recorded exploration around the accuracy--speed tradeoff. Its feedback
retains a locally tested Newton configuration with reduced solver iterations,
the accuracy violations observed in earlier parameter changes, and profiling
evidence pointing to \texttt{mj\_step2}. The root remains \texttt{incomplete},
with an explicit outstanding requirement:

\begin{quote}\small
The target of $\leq$60\% simulation time (40\% speedup) has not been achieved;
the best valid configuration reached only $\sim$80--84\% of reference time.
\end{quote}

This records useful intermediate progress while preserving the need to find
a configuration that satisfies both constraints. The 80--84\% figures refer
to local trials in the history, whereas the 110\% figure above comes from
verification of the source execution's delivered file.

\paragraph{VF continues exploring without producing the required file.}
The VF recipient makes 22 tool calls. It reads the evaluation script and
reference model, benchmarks parameter settings, and checks XML parsing and
simulation flags. Execution ends while it is still exploring configurations,
without writing \texttt{/app/model.xml}. The verifier reports:

\begin{quote}\small
Tuned model file does not exist
\end{quote}

Only the unchanged-reference check passes. The file-existence check fails,
and the correctness and speed checks cannot load the missing model, giving
one of four checks passed.

\paragraph{DENSE carries parameter exploration through to delivery.}
The DENSE recipient makes 25 tool calls. It compares integrators and solver
settings, then tests combinations of \texttt{implicitfast}, the PGS solver,
and disabled contact processing. After refining the iteration count, it checks
candidate settings across ten seeds. Eight iterations give a maximum observed
state difference of $1.23\times10^{-5}$, just above the tolerance; nine reduce
it to $5.66\times10^{-6}$ while retaining approximately 55\% of the reference
simulation time in this local check.

The recipient writes the tuned file on its nineteenth tool call, using:

\begin{quote}\small\ttfamily
\begin{tabular}{@{}l@{}}
<option integrator="implicitfast" solver="PGS"\\
\quad iterations="9" jacobian="auto">\\
\quad <flag contact="disable"/>\\
</option>
\end{tabular}
\end{quote}

It then runs \texttt{eval.py}, repeats the evaluation, and checks that the
reference file is unchanged. The post-execution verifier passes all four
checks and records simulation time at approximately 55\% of the reference.
The retained history supplies intermediate findings and outstanding
constraints; the recipient completes the remaining search, writes the model,
and validates the resulting artifact.

\begin{table}[!htbp]
  \centering
  \caption{MiniMax on \texttt{tune-mjcf}. Checks and tool calls describe the
  executions discussed above; calls are counted from actual transcript
  records. The final column reports strict passes across all three executions
  of this task for each condition.}
  \label{tab:minimax-vf-case}
  \small
  \setlength{\tabcolsep}{8pt}
  \def\minimaxCaseCondition{Condition}
  \def\minimaxCaseChecks{\shortstack{Verifier checks\\passed}}
  \def\minimaxCaseTools{Tool calls}
  \def\minimaxCasePasses{\shortstack{Strict passes\\across the task}}
  \begin{tabular}{lccc}
\toprule
\minimaxCaseCondition & \minimaxCaseChecks & \minimaxCaseTools & \minimaxCasePasses \\
\midrule
Baseline & 3/4 & 44 & 0/3 \\
Self-reflection & 3/4 & 32 & 1/3 \\
\textbf{DENSE (ours)} & \textbf{4/4} & 25 & \textbf{3/3} \\
VF & 1/4 & 22 & 1/3 \\
\bottomrule
\end{tabular}

\end{table}

\paragraph{From exploration to a tested artifact.}
Across the three executions of this task, DENSE passes three, while VF and
Self-reflection each pass one (Table~\ref{tab:minimax-vf-case}). The detailed
comparison shows active parameter exploration under both DENSE and VF.
DENSE's feedback makes partial progress and the remaining target explicit;
the subsequent execution tests new combinations and completes the required
artifact. The case illustrates how trajectory-derived evidence can support
continued optimization through verification and delivery.

\subsection{Complementary Outcomes and Overall Interpretation}
\label{app:feedback-interpretation}

For \texttt{gpt2-codegolf}, the recipient investigates checkpoint offsets and
changes the mapping from logical layers to physical weight blocks to
\texttt{\{0,1,4,5,6,7,8,9,10,11,2,3\}}, together with corrections to embedding
and final LayerNorm offsets. The resulting 3,578-byte C program passes the
verifier. The feedback identifies unverified layout assumptions and repeated
token output; it does not supply this mapping.
Step-by-step also succeeds in this repetition, using 11 tool calls compared
with DENSE's 24. This case illustrates a useful representation of nested
progress, while showing that another feedback view can support a successful
and shorter execution.

\begin{table}[!htbp]
  \centering
  \caption{Three-repeat outcomes. Entries report strict pass rate (\%),
  with strict passes out of three in parentheses. The process analysis uses
  \texttt{password-recovery} repeat 0, \texttt{largest-eigenval} repeat 2, and
  \texttt{gpt2-codegolf} repeat 0. Within each case, bold / underline mark the
  highest / second-highest rates, including ties.}
  \label{tab:case-outcomes}
  \small
  \setlength{\tabcolsep}{4pt}
  \renewcommand{\arraystretch}{1.10}
  \begin{tabular}{@{}lccccc@{}}
\toprule
Task / recipient & Advisor & Self-reflection & All-at-once & Step-by-step & \shortstack{DENSE\\(ours)} \\
\midrule
\shortstack[l]{\texttt{password-recovery}\\Kimi K2.6} & \underline{0.0 (0/3)} & \underline{0.0 (0/3)} & \underline{0.0 (0/3)} & \underline{0.0 (0/3)} & \textbf{66.7 (2/3)} \\
\addlinespace[5pt]
\shortstack[l]{\texttt{largest-eigenval}\\Kimi K2.6} & 0.0 (0/3) & \underline{33.3 (1/3)} & \underline{33.3 (1/3)} & \underline{33.3 (1/3)} & \textbf{100.0 (3/3)} \\
\addlinespace[5pt]
\shortstack[l]{\texttt{gpt2-codegolf}\\GPT-5.5} & \underline{33.3 (1/3)} & 0.0 (0/3) & 0.0 (0/3) & \underline{33.3 (1/3)} & \textbf{66.7 (2/3)} \\
\bottomrule
\end{tabular}

\end{table}

The three trajectory-feedback cases show how historical observations can be organized as
reusable progress and explicit remaining goals, with subsequent actions that
correspond to those goals. The three-repeat results also limit a stronger
interpretation: DENSE fails one \texttt{password-recovery} repetition, and
alternative feedback succeeds in \texttt{gpt2-codegolf}. The five-condition
comparisons jointly change synthesis, hierarchy, length, and prominence; this analysis
does not identify the causal contribution of hierarchy alone. For example,
the concrete concatenated answer in the password feedback is part of the
intervention, not merely a formatting change. A content-matched comparison
would be needed to separate that synthesis from its hierarchical presentation.

The MiniMax \texttt{tune-mjcf} case in Section~\ref{app:minimax-vf} extends this
analysis to privileged feedback: both recipients explore configurations, while
DENSE-guided execution proceeds to a tuned model that passes verification.

\FloatBarrier
\Needspace{6\baselineskip}
\section{Prompts Used in the Experiments}
\label{app:prompts}

This section organizes the prompts by their role in feedback delivery, baseline
extraction, and DENSE construction and reconciliation. The boxes reproduce the
English instructions from the implementation. \textbf{System message} denotes
a complete system message; \textbf{excerpt} denotes a selected passage, with
internal omissions marked explicitly. \textbf{Required output} gives the original
output constraints. The surrounding text describes inputs, invocation points,
and intended behavior. Content slots in braces and output types in angle brackets
are instantiated for each example; line wrapping and indentation are adjusted
for readability. \texttt{Shortcut-Tree} in the original prompts is the
implementation name of DENSE.
Appendix~\ref{app:dense-algorithm} connects these prompts through complete
control flow, structured interfaces, and deterministic validation rules.

Table~\ref{tab:prompt-index} maps each prompt to its input view. Except for the
privileged Verifier Feedback condition, generators do not receive post-hoc run0
rewards, hidden verifier outputs, or reference answers. Public tests already
observed within the trajectory remain admissible evidence under the boundary in
Section~\ref{sec:refit}.

\begin{table}[H]
  \centering
  \caption{Prompt roles and input views. The recipient wrapper is static text;
  the remaining rows identify analysis-model roles.}
  \label{tab:prompt-index}
  \small
  \setlength{\tabcolsep}{4pt}
  \renewcommand{\arraystretch}{1.12}
  \begin{tabular}{p{.07\linewidth}p{.30\linewidth}p{.55\linewidth}}
    \toprule
    Prompt & Role & Visible input \\
    \midrule
    P\ref{prompt:recipient} & Recipient wrapper & Generated feedback, followed by the original task \\
    P\ref{prompt:advisor} & Task-only Advisor & Original task instruction only \\
    P\ref{prompt:all-at-once} & Global action scoring & Task, complete OA sequence, and action indices \\
    P\ref{prompt:step-by-step} & Prefix action scoring & Task and prefix through the target action; its result is hidden \\
    P\ref{prompt:boundary} & Subtask boundaries & Root task, current level, and consecutively numbered queue \\
    P\ref{prompt:local-score} & Local action scoring & Root task, newly formed subtask, target action and observation \\
    P\ref{prompt:summary} & Subtask summaries & Actions, observations, local scores, or child summaries \\
    P\ref{prompt:termination} & Hierarchy termination & Root task and candidate top-level nodes with summaries \\
    P\ref{prompt:cleaner} & Shortcut compression & Ordered direct children of one parent and evidence IDs \\
    P\ref{prompt:critic} & Issue reconciliation & Cleaned subtree, inherited issues, and available evidence IDs \\
    \bottomrule
  \end{tabular}
\end{table}

\subsection{Recipient-Side Feedback Delivery}
\label{app:prompt-recipient}

The run1 task message contains the feedback, a separator, and the original task
instruction, in that order. Prompt~P\ref{prompt:recipient} requires the recipient
to recreate deliverables in the fresh environment and to treat historical paths,
files, and processes as evidence. The slot \texttt{\{content\}} is replaced by
the feedback for the selected condition. This wrapper is used for trajectory
feedback and Verifier Feedback; Advisor instead uses the separate recipient
template in Prompt~P\ref{prompt:advisor}. Run0 receives the original task
instruction without this feedback prefix.

\PromptRecipient

\subsection{Baseline Prompts: Task Priors, Full Trajectories, and Prefixes}
\label{app:prompt-baselines}

\paragraph{Task-only Advisor.}
The advisor receives only the task description and returns a strategy,
instruction-inferable risks, and a verification checklist. Its prompt prohibits
presenting general advice as an error or observation that has already occurred.
The recipient is also told that the advisory was generated without observing a
previous execution.

\PromptAdvisor

\paragraph{All-at-once.}
One call receives the complete OA sequence and the indices of all target actions,
and returns a label and rationale for every action. Later trajectory events may
establish that an earlier action was redundant, undone, or invalidated. This
within-trajectory hindsight does not provide access to the external verifier.
Prompt~P\ref{prompt:all-at-once} includes the scoring rubric and output instructions.

\PromptAllAtOnce

\paragraph{Step-by-step.}
Each action is assessed in a separate call using the prefix ending at the action
itself, excluding its immediate observation and all later events. The
\texttt{known\_basis} field cites evidence already established at that point or
states what should have been checked first. A negative label therefore concerns
the justification of the decision, rather than directly identifying execution
failure.

\PromptStepByStep

Self-reflection supplies the serialized actions and observations for the next-round
agent to summarize past experience and reflect on the previous attempt.
Verifier Feedback also serializes the reward and test output. Neither condition
uses a separate analysis-model prompt to generate advice.

\subsection{DENSE Prompts for Subtask-Tree Construction}
\label{app:prompt-construction}

\paragraph{Contiguous semantic boundaries.}
The boundary prompt identifies a complete sub-phase starting at the queue head;
the same operation is repeated at higher levels. In
Prompt~P\ref{prompt:boundary}, \texttt{action\_index} denotes the current queue
ordinal, rather than the original trajectory action index. Zero selects the head
alone, and $-1$ requests more elements. At the final segment of a level, the user
message explicitly forbids $-1$ because no further elements will arrive.

\PromptBoundary

\paragraph{Scoring before summarization.}
DENSE scores actions within each newly formed local subtask before generating
its summary. The XML input marks the subtask with
\texttt{focus="current-subtask"} and the action with \texttt{focus="TARGET"};
scoring also uses the observation immediately following that action.
Prompt~P\ref{prompt:local-score} grounds the assessment in both the local objective
and the root task, while recognizing necessary exploration, debugging, and
correction. These labels are internal analysis signals, not verifier verdicts,
and are omitted from the final DENSE feedback.

\vspace{-2pt}
\PromptLocalScore

\paragraph{Summarizing reusable evidence.}
Prompt~P\ref{prompt:summary} asks for the shortest observed working path while
retaining the verification that supports it. Reusable failures belong in
\texttt{dead\_ends}, and unverified results belong in \texttt{open\_issues}.
The output fields are \texttt{subtitle}, \texttt{summary}, \texttt{artifacts},
\texttt{final\_state}, \texttt{key\_values}, \texttt{key\_mechanisms},
\texttt{critical\_order}, \texttt{dead\_ends}, and \texttt{open\_issues};
fields without relevant content contain \texttt{none}. Parent summaries also
reconcile conflicting child values and preserve ordering requirements.

\PromptSummary

\paragraph{Stopping the hierarchy construction.}
Prompt~P\ref{prompt:termination} tests whether the candidate nodes form a small,
coherent set of top-level phases. A \texttt{can\_mount\_all=true} response means
that hierarchy construction can stop; it does not assert that the original task
succeeded. Further natural groupings require another aggregation level.

\PromptTermination

\subsection{DENSE Prompts for Compression and Issue Reconciliation}
\label{app:prompt-reconciliation}

\paragraph{Compressing sibling attempts.}
The cleaner receives the ordered direct children of one parent and returns a
complete partition in the same order. A \texttt{keep} group contains one source
node; a \texttt{shortcut} group contains at least two consecutive source nodes.
Each group records the reusable failure, shortest working path, observed local
outcome, and remaining obligation, citing supplied evidence and key-action IDs.
If every attempt fails, the failure and open issue must remain visible.

\PromptCleaner

\Needspace{6\baselineskip}
\paragraph{Reconciling completion and inherited issues.}
The critic reads the cleaned subtree and assesses coherence separately from
completion. It may close an inherited issue only when a later node provides
concrete recovery evidence, with supplied IDs recorded in
\texttt{resolution\_evidence}. The completion-scope instruction in
Prompt~P\ref{prompt:critic} is also supplied to the cleaner: a \texttt{complete}
verdict describes what the trajectory evidence supports.

\PromptCritic

\paragraph{Turning structured outputs into feedback.}
The implementation validates partition coverage, node references, and issue
states before rendering the feedback described in Section~\ref{sec:method}.
Completed branches are folded, unresolved issue paths remain expanded, and
selected key actions retain their original content. Rendering is deterministic
and does not call a model to rewrite the final feedback, so it has no additional
generation prompt.

\clearpage
\section{DENSE Algorithm and Validation Rules}
\label{app:dense-algorithm}

This appendix specifies the control flow and structured interfaces of DENSE.
Algorithms~\ref{alg:dense-pipeline}--\ref{alg:dense-reconcile} connect the prompts
in Appendix~\ref{app:prompts}: P\ref{prompt:boundary}--P\ref{prompt:termination}
build the hierarchy, and P\ref{prompt:cleaner}--P\ref{prompt:critic} compress and
reconcile it. Model judgments determine semantic grouping, useful working paths,
and recovery; deterministic operations maintain source order, check structured
outputs and references, derive states, and render the feedback.

\subsection{Inputs, Outputs, and Execution Order}
\label{app:dense-algorithm-overview}

The inputs are the task $x$ and its visible trajectory
$\tau=(a_1,o_1,\ldots,a_T,o_T)$. Each leaf binds an action to its immediately
following observation, if present. Node IDs identify tree elements; global
action indices identify source actions; queue ordinals identify positions in
the current aggregation level. These three identifiers have distinct roles.
The outputs are a dense source tree $\mathcal T$, a reconciled shortcut tree
$\mathcal T^\star$, and the recipient-visible feedback $f$. Original action
and observation text remains available in the indexed source tree.

\begin{algorithm}[H]
  \caption{DENSE: from a visible trajectory to recipient feedback}
  \label{alg:dense-pipeline}
  \small
  \begin{algorithmic}[1]
\Require Task $x$, trajectory $\tau$, chunk size $\Delta$, round cap $L$, byte budgets $B_t,B_f$
\Ensure Dense source tree $\mathcal T$, reconciled tree $\mathcal T^\star$, recipient feedback $f$
\Function{DENSE}{$x,\tau,\Delta,L,B_t,B_f$}
  \State $Q\gets$ ordered leaves, each binding an action to its following observation
  \If{$|Q|>1$}
    \For{$k=1,\ldots,L$}
      \State $Q'\gets\Call{BuildLevel}{x,Q,k,\Delta}$ \Comment{Algorithm~\ref{alg:dense-boundaries}}
      \If{$|Q'|=1$ or $|Q'|=|Q|$}
        \State $Q\gets Q'$; \textbf{break}
      \EndIf
      \State $s\gets\Call{TopLevelCheck}{x,Q',|Q|}$ \Comment{P\ref{prompt:termination}; false on call/parse failure}
      \State $Q\gets Q'$
      \If{$s=\mathrm{true}$} \State \textbf{break} \EndIf
    \EndFor
  \EndIf
  \State $\mathcal T\gets\Call{Root}{x,Q}$
  \State Score any remaining unscored leaves in singleton local contexts \Comment{P\ref{prompt:local-score}}
  \State Index source nodes and their original action--observation contents
  \State $\mathcal T^\star\gets\Call{Reconcile}{x,\mathcal T}$ \Comment{Algorithm~\ref{alg:dense-reconcile}}
  \State Select recipient-visible key actions from retained evidence \Comment{Section~\ref{app:dense-delivery}}
  \State $f\gets\Call{Render}{\mathcal T^\star,B_t}$ \Comment{States, issues, summaries, and selected actions}
  \State Apply the recipient wrapper and the complete-feedback budget $B_f$
  \State \Return $(\mathcal T,\mathcal T^\star,f)$
\EndFunction
\end{algorithmic}

\end{algorithm}

The configured chunk size is $\Delta=999$ nodes and the maximum number of
aggregation rounds is $L=20$. The tree and complete-feedback budgets are
$B_t=180$~KiB and $B_f=200$~KiB, respectively. P\ref{prompt:local-score} uses
the labels $\{-2,-1,0\}$ internally. Each new multi-node subtask is scored
before its summary is generated; leaves already scored at a lower level keep
their labels. Singletons are promoted unchanged and any still-unscored actions
are scored after hierarchy construction. The pseudocode presents dependencies
sequentially: independent local scores, summaries, and sibling subtrees can
run concurrently. All summaries for a level finish before the next-level
boundary or termination calls read them.

\subsection{Subtask Boundary Decisions}
\label{app:dense-boundary-rules}

\paragraph{Interface.}
P\ref{prompt:boundary} receives the root task, current level, and a consecutively
numbered queue. At the leaf level it reads action text; at higher levels it
reads subtask titles and summaries. Its JSON output contains
\texttt{reasoning} and \texttt{action\_index}. The latter identifies the end of
a coherent prefix beginning at the queue head. A value of $-1$ requests more
queue elements, and zero selects the head alone. The prompt asks for a complete
local objective at level one and progressively broader phases at higher levels.

\begin{algorithm}[H]
  \caption{Ordered prefix grouping and boundary normalization}
  \label{alg:dense-boundaries}
  \small
  \begin{algorithmic}[1]
\Require Task $x$, ordered level $Q$, level index $k$, chunk size $\Delta$
\Ensure Ordered next level $H$, covering every member of $Q$ exactly once
\Function{BuildLevel}{$x,Q,k,\Delta$}
  \State $W\gets[\,]$; $H\gets[\,]$; $h\gets1$ \Comment{$h$ is the current head ordinal in $Q$}
  \For{each consecutive chunk $C$ of at most $\Delta$ nodes from $Q$}
    \State Append $C$ to $W$; $\mathit{last}\gets(C\text{ is the final chunk})$
    \Repeat
      \State $d\gets\Call{Boundary}{x,W,k,h,\mathit{last}}$
      \If{$d=\mathrm{WAIT}$} \State \textbf{break} \EndIf
      \State Remove prefix $G$ of length $d-h+1$ from $W$; $h\gets d+1$
      \If{$|G|=1$}
        \State Append the unchanged node in $G$ to $H$
      \Else
        \State $v\gets\Call{Subtask}{G}$
        \State Score unscored action leaves in the local tree rooted at $v$ \Comment{P\ref{prompt:local-score}}
        \State Populate $v$'s title, summary, and structured facets \Comment{P\ref{prompt:summary}}
        \State Append $v$ to $H$
      \EndIf
    \Until{$\neg\mathit{last}$ or $W=[\,]$} \Comment{One cut per nonfinal chunk; drain the final chunk}
  \EndFor
  \State \Return $H$
\EndFunction
\Statex
\Function{Boundary}{$x,W,k,h,\mathit{last}$}
  \State $t\gets h+|W|-1$
  \State Query P\ref{prompt:boundary} with $x,k$, ordinals $h{:}t$, the queue view, and $\mathit{last}$
  \State Parse \texttt{action\_index} as integer $r$; on call/parse failure \Return $h$
  \If{$r=-1$}
    \If{$\mathit{last}$} \State \Return $t$ \Else \State \Return $\mathrm{WAIT}$ \EndIf
  \EndIf
  \State \Return $\max(h,\min(r,t))$ \Comment{Zero selects the head; normalize out-of-range ordinals}
\EndFunction
\end{algorithmic}

\end{algorithm}

\paragraph{Coverage and stopping.}
Removing only prefixes and appending their replacements in order preserves
every source element exactly once at each level. A singleton introduces no
extra parent. P\ref{prompt:termination} receives the candidate top-level nodes
and returns \texttt{can\_mount\_all}; this is a hierarchy-stopping decision.
Construction also stops if a round leaves the node count unchanged, produces
one node, or reaches the round cap. The remaining nodes are attached to the
task root. Boundary normalization in Algorithm~\ref{alg:dense-boundaries}
ensures progress even when a boundary response cannot be used directly.

\subsection{Shortcut Compression and Issue Reconciliation}
\label{app:dense-reconcile-rules}

Processing is bottom-up: each parent receives its children's already reconciled
representations. Table~\ref{tab:dense-module-interfaces} specifies the two model
interfaces. The full JSON field definitions appear in
P\ref{prompt:cleaner} and P\ref{prompt:critic}.

\begin{table}[H]
  \centering
  \caption{Inputs and outputs at a parent node $v$. Source and evidence IDs always
  refer to supplied trajectory-derived records.}
  \label{tab:dense-module-interfaces}
  \small
  \setlength{\tabcolsep}{5pt}
  \renewcommand{\arraystretch}{1.12}
  \begin{tabular}{L{.14\linewidth}L{.37\linewidth}L{.41\linewidth}}
    \toprule
    Module & Input & Output \\
    \midrule
    Compression (P\ref{prompt:cleaner})
      & Task, parent title/summary, ordered direct children, child verdicts,
        available evidence, internal action-label hints, and deterministic key-action IDs.
      & Ordered groups: mode, source IDs, title, failed approach, working path,
        local outcome, open issue, evidence IDs, and selected key-action IDs. \\
    Reconciliation (P\ref{prompt:critic})
      & Task, cleaned subtree, prior summary, child verdicts, inherited issue IDs,
        and retained evidence/actions.
      & Coherence, completion, final-state summary, open/fatal issues, resolved
        inherited issue IDs, recovery-evidence map, and key-action IDs. \\
    \bottomrule
  \end{tabular}
\end{table}

\begin{algorithm}[H]
  \caption{Bottom-up compression, reconciliation, and validated model calls}
  \label{alg:dense-reconcile}
  \small
  \begin{algorithmic}[1]
\Require Task $x$, source node $v$, indexed original evidence
\Ensure Reconciled node $v^\star$ with source references, retained actions, and validated verdict
\Function{Reconcile}{$x,v$}
  \If{$v$ is a leaf} \State \Return identity copy of $v$ \EndIf
  \State $C\gets[\Call{Reconcile}{x,c}:c\in\operatorname{children}(v)]$ in source order
  \State $K\gets\Call{DeterministicKeys}{C}$; mark these actions for original-content retention
  \State $C'\gets C$
  \If{$|C|\geq2$}
    \State $q\gets\Call{CleanerInput}{x,v,C,K}$ \Comment{P\ref{prompt:cleaner}}
    \State $G\gets\Call{CheckedCall}{q,\operatorname{ValidatePartition}(\cdot,C),\bot}$
    \If{$G\neq\bot$}
      \State $C'\gets\Call{CompileGroups}{G,C,K}$ \Comment{Keep singletons; compile shortcut groups}
    \EndIf
  \EndIf
  \State $u\gets$ root/subtask copy of $v$ with children $C'$ and their source-action indices
  \State $q\gets\Call{CriticInput}{x,u,\operatorname{IssueIDs}(C')}$ \Comment{P\ref{prompt:critic}}
  \State $z\gets\Call{CheckedCall}{q,\operatorname{ValidateVerdict}(\cdot,u),\operatorname{UnknownVerdict}}$
  \State Attach $z$ to $u$; derive $u$'s state by Equation~\ref{eq:dense-state-rule}
  \State Use $z$'s summary if nonempty; otherwise retain $u$'s prior summary
  \State Mark the original actions selected by $z$'s valid key-action IDs
  \State \Return $u$
\EndFunction
\Statex
\Function{CheckedCall}{$q,V,z_{\mathrm{fallback}}$}
  \State $q'\gets q$
  \For{$j=1,2$}
    \State Attempt $r\gets\operatorname{ParseJSON}(\operatorname{Model}(q'))$ and $z\gets V(r)$
    \If{the call, parsing, and validation succeed} \State \Return $z$ \EndIf
    \State $q'\gets q+$ previous response $+$ validation error
    \State Append the instruction to return a corrected JSON object
  \EndFor
  \State \Return $z_{\mathrm{fallback}}$
\EndFunction
\end{algorithmic}

\end{algorithm}

\paragraph{Compiling a shortcut.}
A \texttt{keep} group retains its single source node. A \texttt{shortcut}
group replaces at least two consecutive siblings, recording their node IDs,
source-action indices, failed approach, working path, local outcome, and
remaining issue. It also carries inherited issue IDs from those sources.
Key actions are copied from source records, including their observations.
The retained set combines model-selected IDs with deterministic keys: recognized
write/edit operations, shell operations that write or change files or run
tests, and the last action under each parent. Thus, choosing a shortcut does
not ask the model to regenerate executable action text.

\paragraph{Reconciling an issue.}
An issue is represented by its description, evidence references, and a unique
identifier linking it to subsequent recovery evidence.
P\ref{prompt:critic} assesses whether a later cleaned node provides concrete
recovery evidence for an inherited issue. The response names that issue in
\texttt{resolved\_issue\_ids} and supplies its evidence in
\texttt{resolution\_evidence}. The checks below validate those identifiers
before accepting the update. The renderer collects accepted resolved IDs and
omits matching inherited issue entries. Local shortcut outcome and parent
completion remain separate fields: a recovered path can support a still
unfinished parent task.

\subsection{Validation Rules and State Derivation}
\label{app:dense-validation}

For an ordered child list $C=(c_1,\ldots,c_m)$, let $R(C)$ be the supplied
node, source, and evidence references reachable from that representation,
$A(C)$ its retained action-node IDs, and $I(C)$ its inherited issue IDs.
Validation checks each reference against these supplied sets.

\begin{table}[H]
  \centering
  \caption{Deterministic checks on compression and reconciliation outputs.}
  \label{tab:dense-validation}
  \small
  \setlength{\tabcolsep}{5pt}
  \renewcommand{\arraystretch}{1.14}
  \begin{tabular}{L{.23\linewidth}L{.69\linewidth}}
    \toprule
    Checked object & Acceptance rule \\
    \midrule
    Sibling partition
      & Concatenating all \texttt{source\_node\_ids} lists must equal
        $(c_1,\ldots,c_m)$ exactly. Groups are nonempty, contiguous, and ordered;
        \texttt{keep} has one source and \texttt{shortcut} has at least two. \\
    Shortcut fields
      & Mode is \texttt{keep} or \texttt{shortcut}; outcome is
        \texttt{succeeded}, \texttt{failed}, \texttt{partial}, or \texttt{unknown}.
        Every cited evidence ID belongs to $R(C)$. \\
    Key actions
      & Selected IDs belong to $A(C)$; cleaner-selected actions also belong to
        their own source group. Compilation unions these IDs with deterministic
        keys and copies original records. \\
    Critic fields
      & Coherence is \texttt{coherent}, \texttt{inconsistent}, or
        \texttt{insufficient\_evidence}; completion is \texttt{complete},
        \texttt{incomplete}, \texttt{failed}, or \texttt{unknown}. \\
    New issues
      & Issue lists contain nonempty descriptions; every supplied evidence ID
        is valid. Each issue has a unique identifier. \\
    Resolved issues
      & Every resolved ID belongs to $I(C')$. Each has a nonempty recovery-evidence
        list, all of whose references are valid in the cleaned subtree. \\
    \bottomrule
  \end{tabular}
\end{table}

Let $c$ and $p$ denote validated coherence and completion, and let
$O$ and $F$ be the reported open- and fatal-issue lists. The final state is
computed in the following priority order:
\begingroup\small
\begin{equation}
\operatorname{state}(c,p,O,F)=
\begin{cases}
\texttt{broken}, & c=\texttt{inconsistent}\ \lor\ p=\texttt{failed}\ \lor\ F\neq\emptyset,\\
\texttt{incomplete}, & p=\texttt{incomplete}\ \lor\ O\neq\emptyset,\\
\texttt{unknown}, & c=\texttt{insufficient\_evidence}\ \lor\ p=\texttt{unknown},\\
\texttt{complete}, & \text{otherwise}.
\end{cases}
\label{eq:dense-state-rule}
\end{equation}
\endgroup
The first satisfied condition applies. The model supplies the evidence-based
judgments; the program computes this state from validated fields. Both cleaner
and critic are instructed to ground success in observations, verification
visible in the trajectory, or completed child verdicts. Numeric action labels
serve as diagnostic hints. Under the DENSE information boundary, the root
completion scope is \texttt{trajectory\_local}.

\paragraph{Handling an invalid response.}
Compression and reconciliation each allow one corrective call with the original
input, the previous response, and the validation error. If both attempts fail,
compression retains the original child list; reconciliation retains the prior
summary and returns an \texttt{insufficient\_evidence}/\texttt{unknown} verdict.
Algorithm~\ref{alg:dense-reconcile} specifies these handling rules.

\paragraph{Bounded module views.}
Module inputs use length-limited views of actions, observations, and summaries.
Original action--observation records remain available for feedback rendering.

\subsection{Selecting Evidence and Producing Feedback}
\label{app:dense-delivery}

\paragraph{Recipient-visible evidence.}
From the retained actions, the deterministic selector includes critic-selected
actions and available action references supporting unresolved issues or their
resolution. It also selects branch representatives for product creation,
verification, final responses, and terminal actions. Shortcuts retain available
failure and working-path representatives; their reusable failure/recovery
lessons are collected at the root. Selection introduces no additional model call.

Required evidence representatives are selected first. Additional retained actions
are prioritized by their roles in delivery, verification, and model-selected
evidence, subject to an adaptive byte budget. The renderer applies the overall
tree budget $B_t$ to the final feedback.

\paragraph{State-guided rendering and budget handling.}
The initial rendering presents the root summary, scope, and unresolved issues.
Subtasks marked \texttt{complete} are folded to summaries and
selected key actions; \texttt{incomplete}, \texttt{broken}, and \texttt{unknown}
subtasks expand recursively. Numeric action labels are omitted. If the result
exceeds $B_t$, the renderer first condenses succeeded shortcuts, then folds
non-root subtask detail while retaining the top-level overview, issues, and
selected actions. If necessary, it removes entire selected-action blocks in
reverse traversal order. A final compact root-and-scope record handles any
remaining overflow. Retained key-action bodies are serialized from source text
without model rewriting. The complete-feedback budget is then applied, retaining
the structured tree while dropping optional surrounding sections as whole
blocks. Appendix~\ref{app:feedback-structure} shows the resulting interface.

\Needspace{6\baselineskip}
\section{Full-Result Visualizations}
\label{app:full-result-figures}

Table~\ref{tab:full-outcomes} reports strict pass rates and their paired changes;
Table~\ref{tab:full-tokens} reports recipient token use and changes from the
baseline row. The figures below show the same results across methods and tasks.
All methods use a fixed order, with the initial execution as Baseline and
Verifier Feedback (VF) distinguished by its additional outcome information.

Figure~\ref{fig:full-token-bars} reports execution-token changes.
Figure~\ref{fig:absolute-token-cost} shows estimated USD costs for GPT-5.5 execution
and Gemini feedback, using the prices in Appendix~\ref{app:feedback-cost}.
Figure~\ref{fig:cost-strict-pass} retains token units
and compares Baseline, Advisor, Self-reflection, All-at-once, Step-by-step, and DENSE
across all four recipients, with equal task weights. Token totals combine
execution and feedback generation per result, following the usage accounting
and cost attribution in Appendix~\ref{app:exp-details}.
Scatter points are model--method aggregates with independent per-model axes and
no fitted trend: left uses fewer tokens, up is better.

\Needspace{.9\textheight}
\subsection{Aggregate Task Performance}
\label{app:full-results-performance}

\begin{figure}[H]
  \small
  \vspace{-4pt}
  \centering
  \includegraphics[width=.84\linewidth]{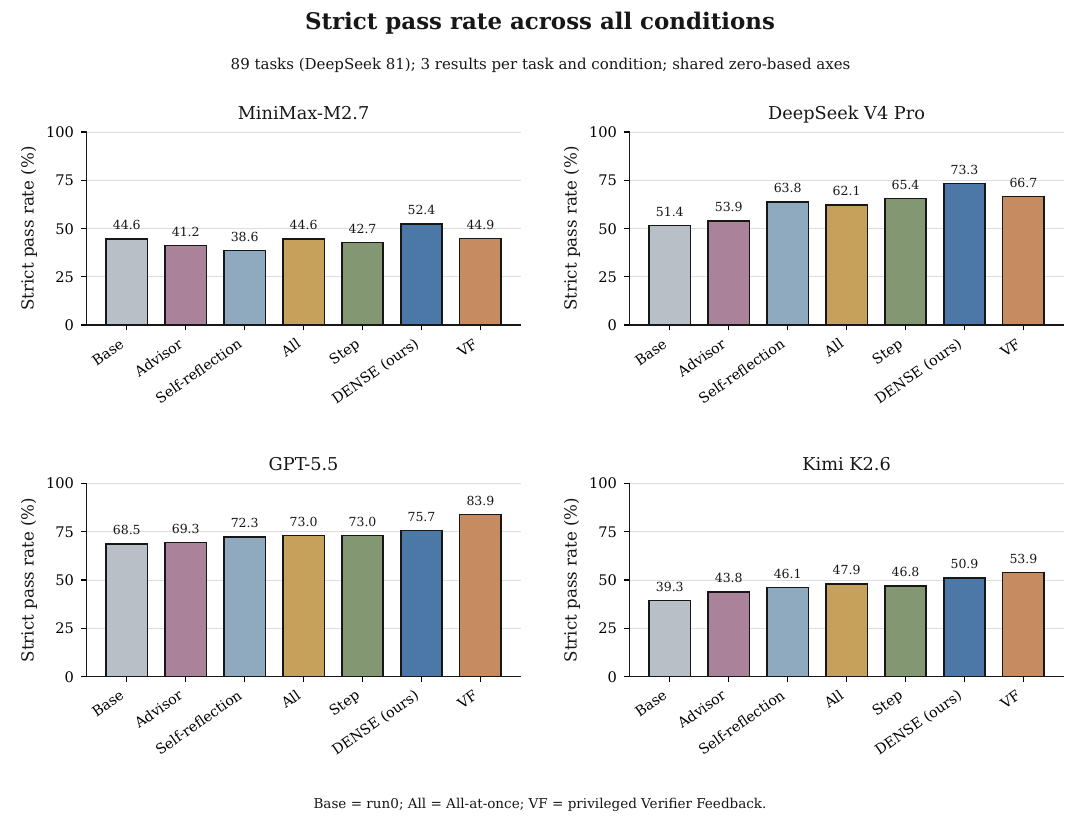}
  \caption{Strict pass rates across all seven conditions, using the outcomes in
  Table~\ref{tab:full-outcomes}. Strict pass requires reward exactly one; rates
  average three repetitions within each task and then weight tasks equally.
  MiniMax, GPT, and Kimi use 89 tasks each, and DeepSeek uses 81. Shared
  zero-based axes show absolute performance on the same percentage scale.
  DENSE leads the tested non-privileged methods for every recipient. VF uses
  additional outcome information and exceeds DENSE on GPT and Kimi, but not on
  MiniMax or DeepSeek, so it serves as a privileged reference rather than a
  guaranteed upper bound.}
  \label{fig:full-zero-outcomes}
  \vspace{-5pt}
\end{figure}
\begin{figure}[H]
  \small
  \vspace{-4pt}
  \centering
  \includegraphics[width=.84\linewidth]{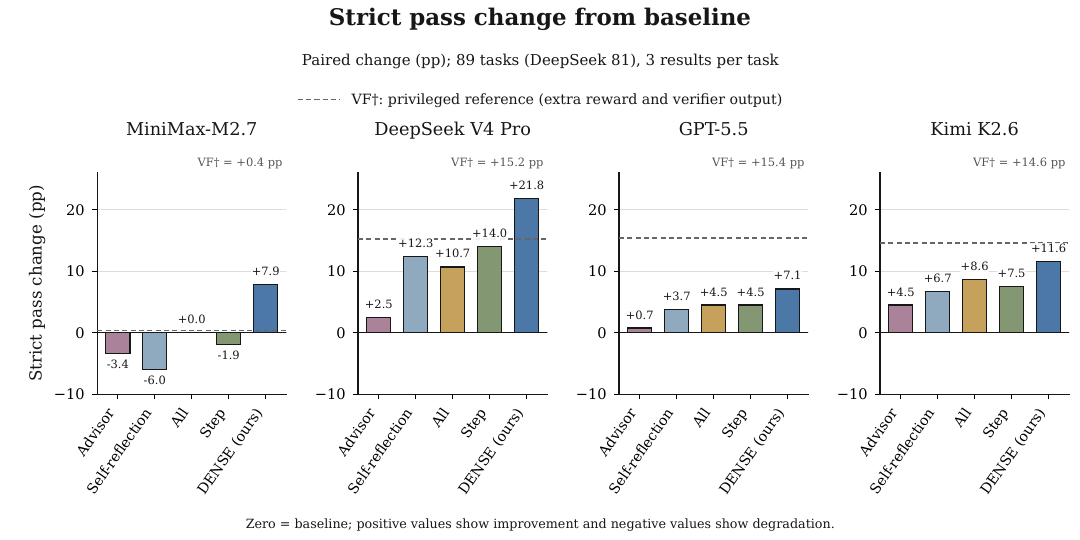}
  \caption{Paired strict pass change from the initial execution, using
  Table~\ref{tab:full-outcomes}. Each value averages the run1-minus-run0 binary
  success difference over matched repetitions and equally weighted tasks,
  expressed in percentage points rather than relative percentages. Zero marks
  unchanged performance; negative bars indicate degradation. Bars show methods
  without post-hoc verifier information, and gray dashed lines show VF.
  DENSE gains 7.87, 21.81, 7.12, and 11.61 pp on MiniMax, DeepSeek, GPT, and
  Kimi, respectively. On MiniMax, it is the only non-privileged feedback method
  with a positive change.}
  \label{fig:full-baseline-gains}
  \vspace{-5pt}
\end{figure}
\Needspace{.9\textheight}
\subsection{Execution Costs, Token Use, and Performance}
\label{app:full-results-cost}

\begingroup
\setlength{\intextsep}{4pt}
\setlength{\abovecaptionskip}{4pt}
\begin{figure}[H]
  \small
  \vspace{-4pt}
  \centering
  \includegraphics[width=.80\linewidth]{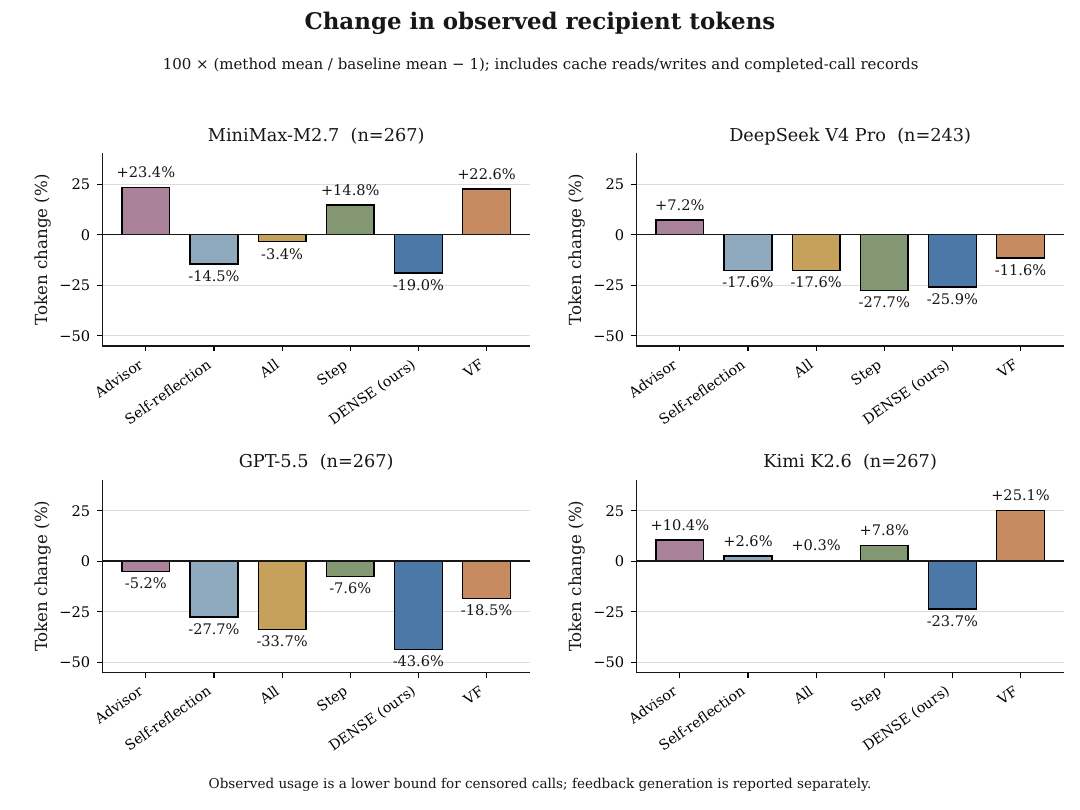}
  \caption{Observed recipient-token change, computed from
  Table~\ref{tab:full-tokens}, excluding feedback generation and following
  Appendix~\ref{app:recipient-token-accounting}. Negative bars indicate lower
  execution use. DENSE reduces it by 19.0--43.6\% across recipients, while
  Step-by-step achieves the largest reduction on DeepSeek.}
  \label{fig:full-token-bars}
  \vspace{-5pt}
\end{figure}
\begin{figure}[H]
  \small
  \vspace{-4pt}
  \centering
  \includegraphics[width=.80\linewidth]{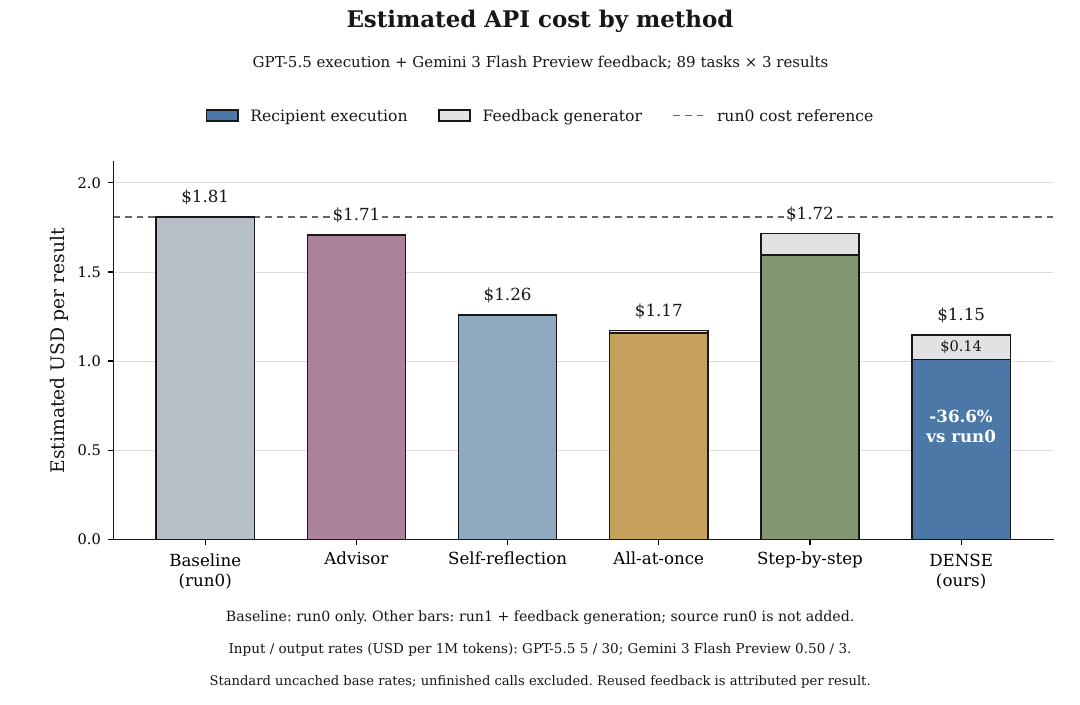}
  \caption{Estimated USD per result for GPT-5.5, using the recorded input/output
  usage and fixed model-specific prices in Appendix~\ref{app:feedback-cost}.
  Costs average three repetitions within each of 89 tasks, then weight tasks
  equally. Method-colored segments denote recipient execution; gray segments
  denote Gemini feedback generation. Baseline and the dashed line show run0;
  other bars combine run1 and feedback.
  DENSE totals \$\gptDenseTotalCost{}, including
  \$\gptDenseFeedbackCost{} for feedback, versus \$\gptBaselineCost{} for run0
  (Table~\ref{tab:gpt55-priced-costs}). The \gptDenseTotalSavingPct\% reduction
  includes feedback generation under this pricing configuration;
  attribution follows Appendix~\ref{app:feedback-attribution}.}
  \label{fig:absolute-token-cost}
  \vspace{-5pt}
\end{figure}
\endgroup
\begin{figure}[H]
  \small
  \vspace{-4pt}
  \centering
  \includegraphics[width=\linewidth]{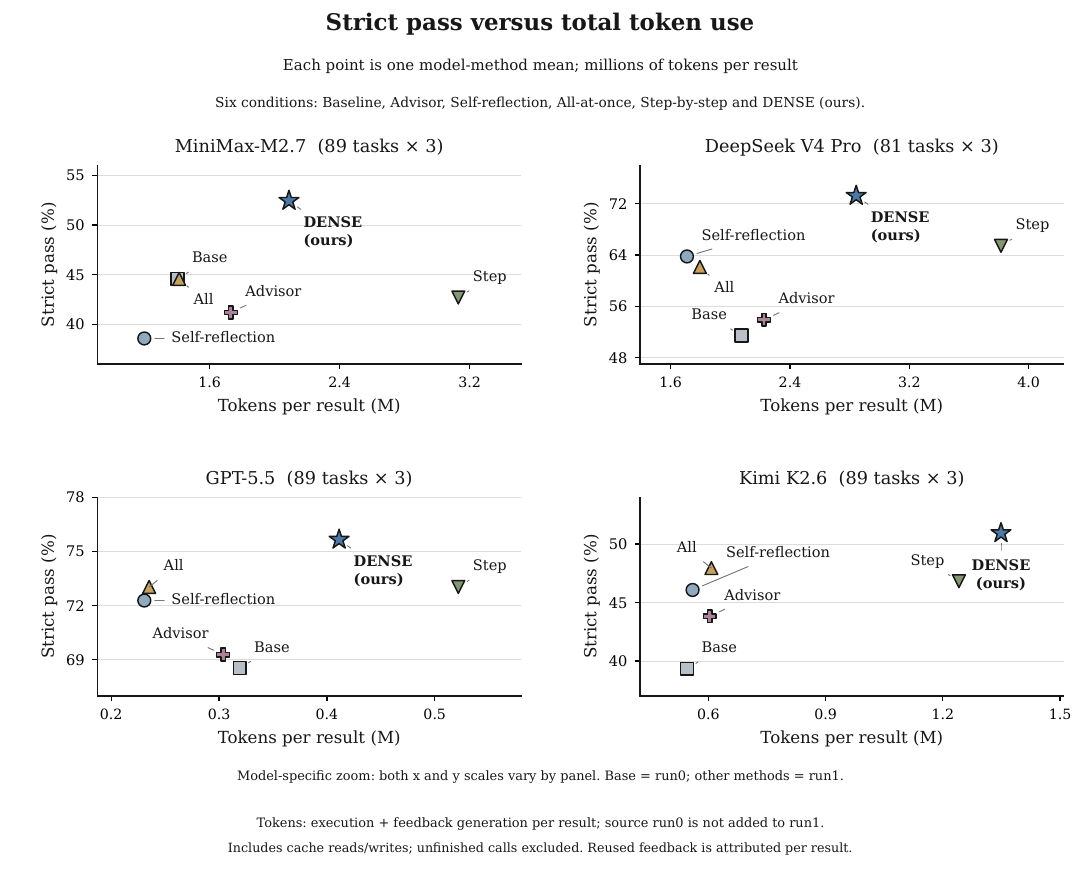}
  \caption{Strict pass rate versus observed execution plus feedback-generation
  tokens per result. Each point aggregates one model--method condition over
  three repetitions per task with equal task weights; VF is excluded.
  Horizontal values sum recipient and attributed generator tokens in millions,
  while vertical values match Table~\ref{tab:full-outcomes}. Accounting follows
  Appendix~\ref{app:exp-details}. Upper-left positions combine higher success with fewer tokens,
  but panel scales differ. DENSE has the highest pass rate in every panel,
  while its combined token use exceeds Baseline for all four recipients.
  Including generation thus exposes overhead absent from the execution-only
  comparison in Figure~\ref{fig:performance-tokens}; token totals do not apply
  the model-specific dollar prices used in Figure~\ref{fig:absolute-token-cost}.}
  \label{fig:cost-strict-pass}
  \vspace{-5pt}
\end{figure}

\clearpage
\subsection{Task-Level Results}
\label{app:full-results-tasks}

Each task heatmap cell reports strict pass rate over three repetitions:
0, 33.3, 66.7, or 100\%. Gray N/A cells are excluded from DeepSeek's denominator.

\begin{figure}[H]
  \centering
  \includegraphics[width=.91\linewidth,height=.84\textheight,keepaspectratio]{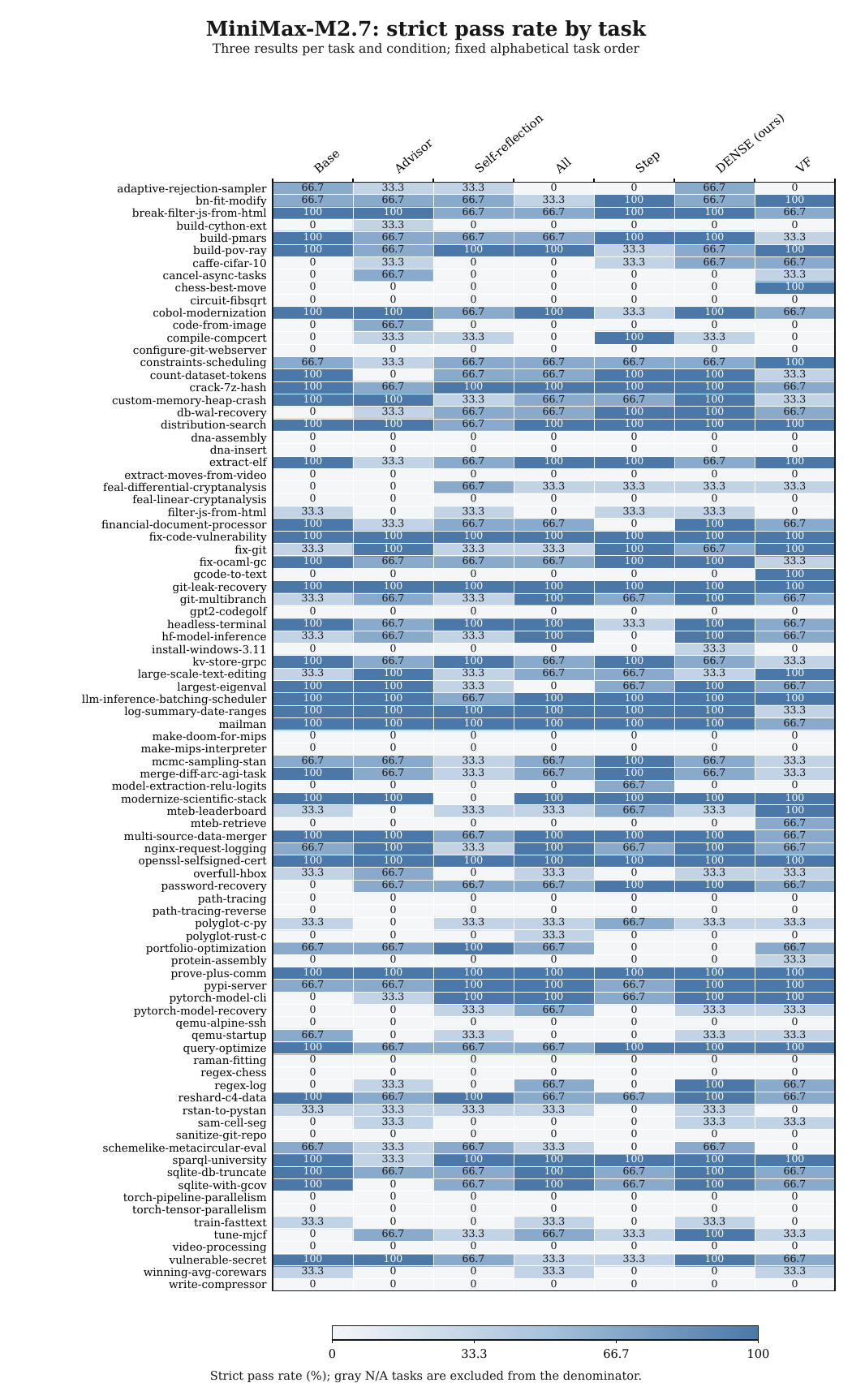}
  \caption{MiniMax-M2.7: all 89 tasks and seven conditions, three results per
  cell. Each cell reports strict pass rate (\%) over three repetitions.}
  \label{fig:tasks-minimax}
\end{figure}
\clearpage
\begin{figure}[H]
  \centering
  \includegraphics[width=.91\linewidth,height=.84\textheight,keepaspectratio]{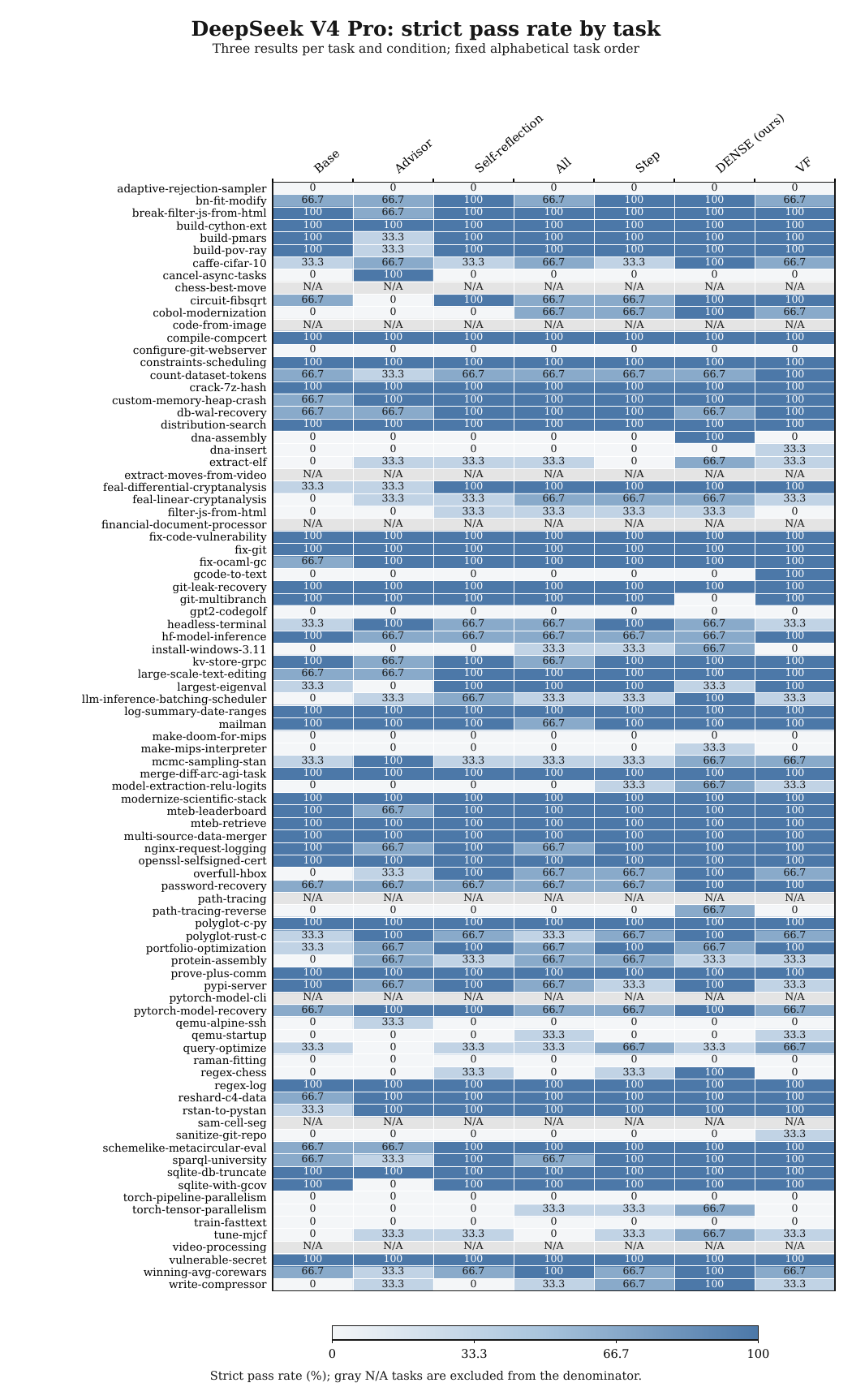}
  \caption{DeepSeek V4 Pro: 81 applicable tasks and eight visual tasks marked
  N/A. Each cell reports strict pass rate (\%) over three repetitions.}
  \label{fig:tasks-deepseek}
\end{figure}
\clearpage
\begin{figure}[H]
  \centering
  \includegraphics[width=.91\linewidth,height=.84\textheight,keepaspectratio]{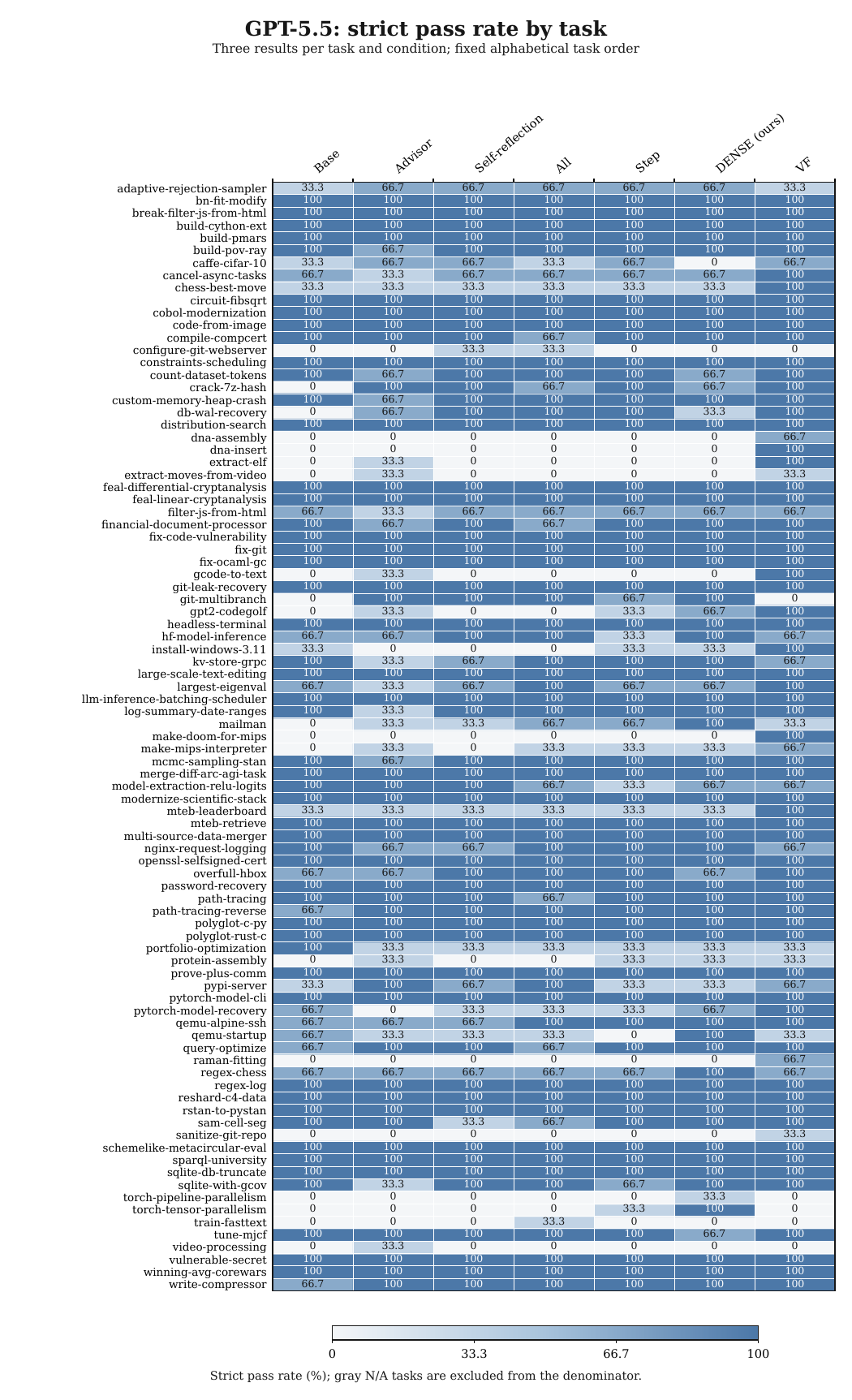}
  \caption{GPT-5.5: all 89 tasks and seven conditions, three results per cell.
  Each cell reports strict pass rate (\%) over three repetitions.}
  \label{fig:tasks-gpt}
\end{figure}
\clearpage
\begin{figure}[H]
  \centering
  \includegraphics[width=.91\linewidth,height=.84\textheight,keepaspectratio]{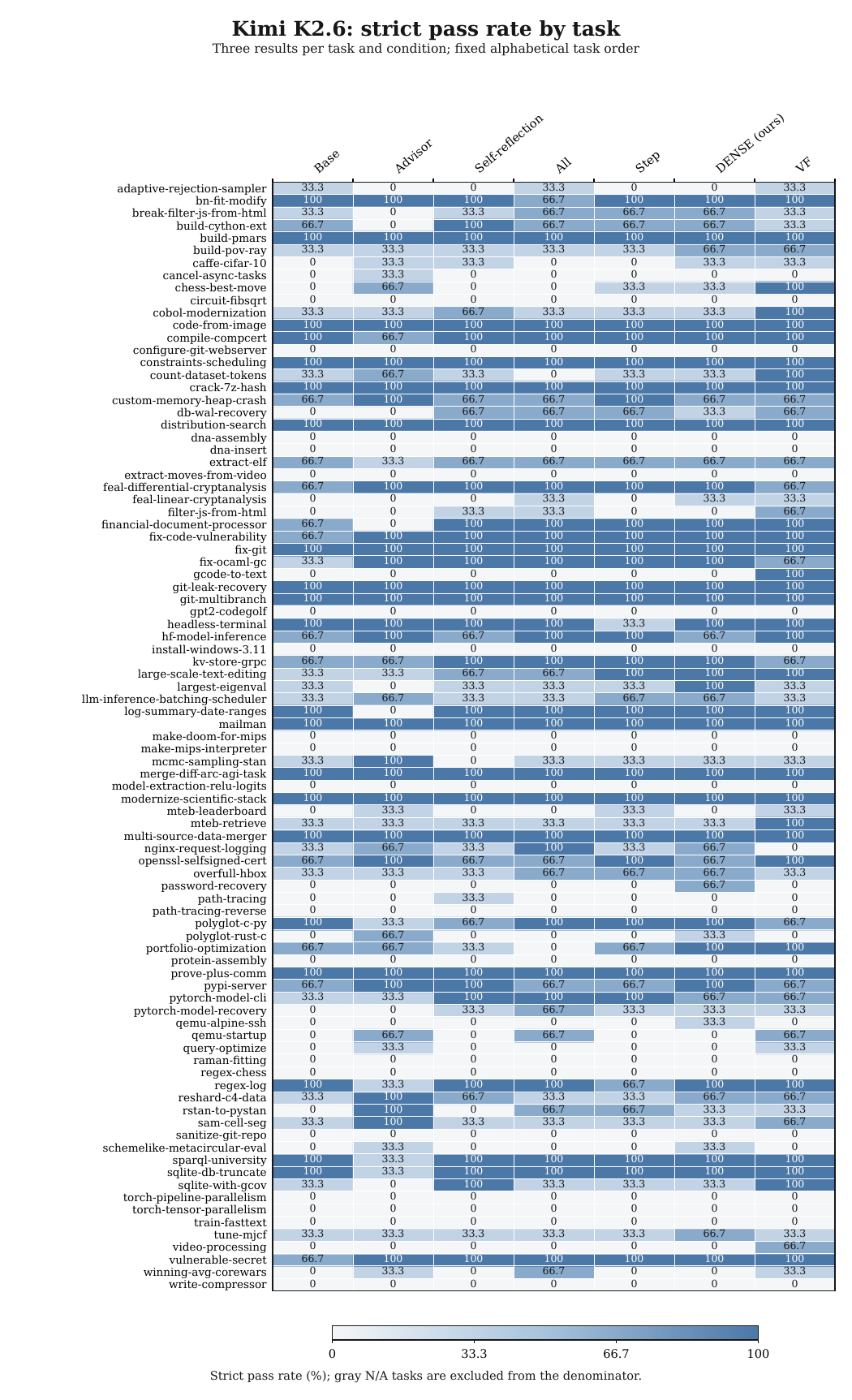}
  \caption{Kimi K2.6: all 89 tasks and seven conditions, three results per cell.
  Each cell reports strict pass rate (\%) over three repetitions.}
  \label{fig:tasks-kimi}
\end{figure}

\end{document}